\documentclass[sigconf, screen, nonacm]{acmart}
\AtBeginDocument{%
  }
\renewcommand\footnotetextcopyrightpermission[1]{}
\usepackage{algorithm}
\usepackage{algpseudocode}
\usepackage{booktabs}
\usepackage{siunitx}
\usepackage{multirow}
\usepackage{enumitem}
\usepackage{graphicx}
\usepackage{xcolor}
\usepackage{amsmath}

\usepackage{tcolorbox}
\usepackage{pifont}
\definecolor{mydarkblue}{RGB}{0, 0, 150} 

\newcommand{\mRatk}{\mathrm{mR}\,@k}
\newcommand{\mIoUatk}{\mathrm{mIoU}\,@k}

\begin{document}

\ccsdesc[500]{Information systems~Retrieval tasks and goals}
\ccsdesc[300]{Computing methodologies~Computer vision}

\title{Retrieving Any Relevant Moments: Benchmark and Models for Generalized Moment Retrieval}
\author{%
Yiming Ding$^{1,2}$ \quad
Siyu Cao$^{1}$ \quad
Luyuan Jiao$^{3}$ \quad
Yixuan Li$^{1}$\\[2pt]
Zitong Wang$^{4}$ \quad
Zhiyong Liu$^{1}$ \quad
Lu Zhang$^{1,*}$\\[3pt]
\small $^{1}$Institute of Automation, Chinese Academy of Sciences
\quad
$^{2}$Beijing University of Posts and Telecommunications\\
\small $^{3}$Wuhan University \quad
$^{4}$University of Electronic Science and Technology of China\\[2pt]
\small \textbf{Code and dataset:} \url{https://github.com/dymm9977/generalized-moment-retrieval}
}

\renewcommand{\shortauthors}{Ding et al.}

\begin{abstract}
Video Moment Retrieval (VMR) aims to localize temporal segments in videos that correspond to a natural language query, but typically assumes only a single matching moment for each query. This assumption does not always hold in real-world scenarios, where queries may correspond to multiple or no moments. Thus, we formulate Generalized Moment Retrieval (GMR), a unified setting that requires retrieving the complete set of relevant moments or predicting an empty set. To enable systematic study of GMR, we introduce Soccer-GMR, a large-scale benchmark built on challenging soccer videos that reflect general GMR scenarios, with realistic negative and positive queries. The benchmark is constructed via a duration-flexible semi-automated pipeline with human verification, enabling scalable data generation while maintaining high annotation quality. We further design a unified evaluation protocol with complementary metrics tailored for null-set rejection, positive-query localization, and end-to-end GMR performance. Finally, we establish strong baselines across two modeling paradigms: a lightweight plug-and-play GMR adapter for discriminative VMR models, and a GMR-tailored GRPO reward for fine-tuning multimodal large language models (MLLMs). Extensive experiments show consistent gains across all metrics and expose key limitations of current methods, positioning GMR as a more realistic and challenging benchmark for video-language understanding.
\end{abstract}

\keywords{video moment retrieval, temporal grounding, benchmark, multi-modal learning}

\maketitle
\begingroup
\renewcommand{\thefootnote}{*}
\footnotetext{Corresponding author.}
\endgroup

\section{Introduction}
Temporally localizing semantic moments is a core capability in video understanding. Video Moment Retrieval (VMR) formalizes this capability as the task of identifying temporal segments in videos that correspond to a natural language query~\cite{zhang2023temporal}. By establishing such cross-modal correspondence, VMR facilitates a wide range of downstream applications, including video question answering~\cite{bai2025qwen3,zhang2025videollama,zhang2024llava}, video dialog~\cite{chen2025grounded,abdessaied2025v,abdessaied2024multi}, multimodal retrieval~\cite{zhang2025bridging,lee2025generalized,xing2025context}, and grounded video reasoning~\cite{deng2025motion,liu2025commonsense,chen2025cross}.

However, existing VMR tasks typically rely on an implicit yet restrictive assumption: each query corresponds to \textit{exactly one} segment in the video. This assumption fundamentally shapes the design of existing datasets, evaluation protocols, and model training objectives~\cite{chen2024verified,liang2025tvr,qin2025generalized}. But in practice, a query may correspond to \textit{multiple or no} relevant moments within a video, requiring models to both retrieve all valid moments and correctly reject queries without corresponding moments. For instance, in a soccer match video, a query like "a corner kick" can occur multiple times, whereas "a red card" or "the goalkeeper saves a penalty kick" may not be present at all. This mismatch between formulation and real-world scenarios poses a fundamental challenge to existing VMR methods~\cite{cao2025one}.

\begin{figure*}[h]
    \centering
    \includegraphics[width=\textwidth]{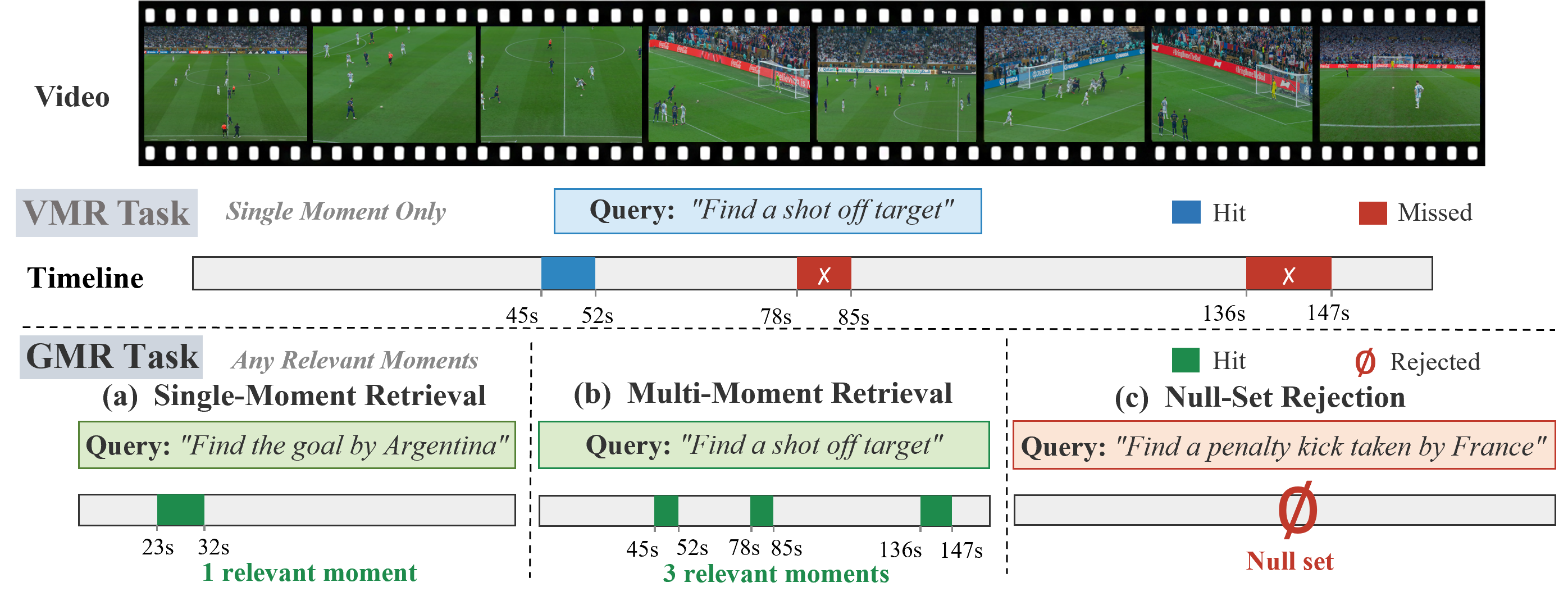}
    \caption{\textbf{Three retrieval scenarios in Generalized Moment Retrieval (GMR).} Given a video and a natural language query, the target moment set may contain (a) exactly one, (b) multiple, or (c) no relevant moments. GMR requires models to localize all matching moments or reject queries when no corresponding moments exist.}
    \Description{Illustration of three GMR scenarios: single-moment retrieval, multi-moment retrieval, and null-set rejection.}
    \label{fig:intro}
  \end{figure*}

To bridge this gap, we consider a more general formulation of the problem, termed \textbf{Generalized Moment Retrieval (GMR)}, where a model is required to return the complete set \textit{(one, multiple, or none)} of temporal segments in a video that correspond to a given natural language query. By this definition, GMR subsumes conventional VMR as a special case while introducing two new challenges: 1) multi-moment retrieval, requiring the model to localize all relevant moments rather than a single best candidate, and 2) null-set rejection, requiring to return an empty set when the queried event is absent. Figure~\ref{fig:intro} illustrates three representative cases of the GMR setting. While prior efforts have attempted to tackle these challenges, they are not yet fully aligned with the GMR setting in three aspects. First, negative samples are predominantly generated by pairing queries with unrelated videos or by randomly modifying key entities (e.g., subject, object, or predicate) to break their semantic alignment with the video~\cite{qin2025generalized,moon2023query}, resulting in queries that are unlikely to arise in real retrieval scenarios, and thus substantially underestimating the difficulty of rejection~\cite{yang2024new}. Second, existing metrics are largely inherited from the conventional VMR task and are not well suited to evaluate models on multiple or absent relevant moments~\cite{flanagan2025moment,qin2025generalized,li2022compositional}. Third, prior works mainly focus on isolated aspects of GMR~\cite{cao2025one,chen2025prvr,flanagan2025moment}, thus lacking a unified framework encompassing data, evaluation, and methods.

To address these challenges, we present a comprehensive study of generalized moment retrieval. First, we introduce a new benchmark named Soccer-GMR, which is instantiated on challenging soccer videos while reflecting general GMR scenarios. The benchmark comprises 5.5k video clips of 139 diverse matches and provides 22.1k query-moment pairs spanning null-set, single-moment, and multi-moment scenarios. We build the benchmark with a duration-flexible semi-automated pipeline that generates structured queries from raw timestamps and caption annotations, producing multi-scale clips with balanced positive and realistic in-domain negative samples. The resulting annotations are further carefully verified by human annotators and experts to ensure quality and consistency. Then, we design a unified evaluation protocol with complementary metrics for null-set rejection, positive-query localization, and overall end-to-end GMR performance.

Finally, we propose GMR-aware methods along two primary paradigms to establish strong baselines. For discriminative VMR models (e.g., DETR-based approaches)~\cite{lei2021detecting,ma2025ms,zhao2025ld}, we propose a lightweight GMR adapter that attaches a parallel existence-estimation branch, enabling explicit null-set prediction without modifying the backbone architecture. For generative MLLM methods, we design a GMR-tailored reward for GRPO-based fine-tuning~\cite{shao2024deepseekmath}, jointly optimizing localization quality and null-set rejection. Extensive experiments on Soccer-GMR demonstrate consistent gains across all metrics while highlighting generalized moment retrieval and temporal localization with MLLMs as key remaining challenges.

Our main contributions are as follows:
\begin{enumerate}[leftmargin=*, nosep, label=\arabic*.]
  \item We introduce Soccer-GMR, a large-scale GMR benchmark comprising 5.5K clips of 139 diverse matches, and 22.1K query-moment pairs with naturally occurring in-domain negatives of high semantic similarity, constructed via a duration-flexible semi-automated pipeline.
  \item To enable systematic evaluation of GMR, we design a unified protocol with metrics for null-set rejection, single-moment and multi-moment retrieval, addressing the gap left by conventional VMR measures.
  \item We propose the GMR Adapter, a lightweight module compatible with mainstream VMR backbones, and design a GMR-tailored reward for GRPO-based fine-tuning on MLLMs. Experiments show that the proposed methods outperform existing baselines, while also exposing open challenges inherent to GMR.
\end{enumerate}

\section{Related Work}
\subsection{Video Moment Retrieval}
Video moment retrieval (VMR) aims to localize temporal segments in a video that correspond to a natural language query~\cite{liu2023survey}. Early proposal-based methods generate candidate segments via sliding windows or predefined anchors and rank them against the query~\cite{lan2023survey}. Proposal-free approaches instead regress boundaries directly from frame-level representations~\cite{woo2024let}. Recently, DETR-based set prediction has become the dominant paradigm. Moment-DETR first introduces learnable query slots with Hungarian matching for parallel moment prediction~\cite{lei2021detecting,carion2020end}, followed by refinements in query-dependency modeling (QD-DETR~\cite{moon2023query}), event-aware slot attention (EaTR~\cite{jang2023knowing}), and correlation-guided cross-attention (CG-DETR~\cite{moon2023correlation}). FlashVTG~\cite{cao2025flashvtg} offers an alternative via multi-scale temporal feature layering without a DETR decoder, achieving competitive performance.

Despite architectural diversity, existing VMR methods share two key limitations. First, they lack an explicit mechanism for null-set rejection: their moment
retrieval objectives (e.g., Hungarian matching with span regression) are
designed for positive query-video pairs and produce no gradient signal
when the queried event is absent, leaving models unable to reject
queries without corresponding moments~\cite{lei2021detecting,moon2023query}.
Second, although set-prediction architectures can in principle output
multiple candidates, the prevailing datasets, evaluation protocols, and task
formulations predominantly assume a single corresponding moment per query,
leaving multi-moment retrieval capacity largely unexploited~\cite{lei2021detecting,gao2017tall}.
Recent multimodal large language models applied to temporal
grounding~\cite{wang2026spacevllm,wu2025survey,pramanick2025enrich} similarly default to single-moment
outputs and exhibit limited fine-grained temporal localization ability.

\subsection{Towards Generalized Moment Retrieval}

The limitations identified above have motivated recent efforts along two
complementary directions.
On the null-set rejection side, Fang et al.~\cite{fang2024not} formalize
Open-Set VMR, treating video-irrelevant queries as an out-of-distribution
detection problem via normalizing flows, while
Flanagan et al.~\cite{flanagan2025moment} propose Negative-Aware VMR, 
distinguishing in-domain from out-of-domain negatives and
benchmarking rejection on existing VMR datasets.
However, negative queries in these works are predominantly constructed via
cross-domain sampling or random entity replacement. Even where in-domain
negatives are considered, they are synthetically generated rather than
naturally occurring, yielding rejection tasks considerably easier than in
realistic in-domain settings.
Moreover, positive queries in both works remain restricted to the
single-moment setting, leaving multi-moment retrieval unaddressed.

On the multi-moment side, Cao et al.~\cite{cao2025one} introduce
Multi-Moment Retrieval (MMR) with the QV-M$^2$ dataset and a
cross-moment post-verification module (FlashMMR), though their formulation
assumes at least one corresponding moment ($n \geq 1$) and does not address
null-set queries.
Qin et al.~\cite{qin2025generalized} propose Generalized VMR (GVMR), the
closest prior formulation to ours, extending VMR to one-to-multi and no-target scenarios with the NExT-VMR benchmark.
While GVMR covers all three scenarios, its negative samples similarly rely
on synthetic construction, and its evaluation protocol inherits
conventional VMR metrics without dedicated measures for generalized retrieval.

\subsection{VMR Benchmarks}
Existing VMR benchmarks, including Charades-STA~\cite{gao2017tall},
ActivityNet Captions~\cite{krishna2017dense},
TACoS~\cite{regneri2013grounding}, and
QVHighlights~\cite{lei2021detecting}, predominantly provide single-moment annotations and lack null-set samples. Recent benchmarks have begun to move beyond this setting: QV-M$^2$~\cite{cao2025one} provides multi-moment annotations but does not address null-set queries, while NExT-VMR~\cite{qin2025generalized} covers both scenarios but lacks evaluation metrics designed for generalized retrieval. Moreover, both are built on short clips with durations fixed at construction time, limiting their applicability to long-form video retrieval research. 

\begin{figure*}[h]
  \centering
  \includegraphics[width=0.95\textwidth]{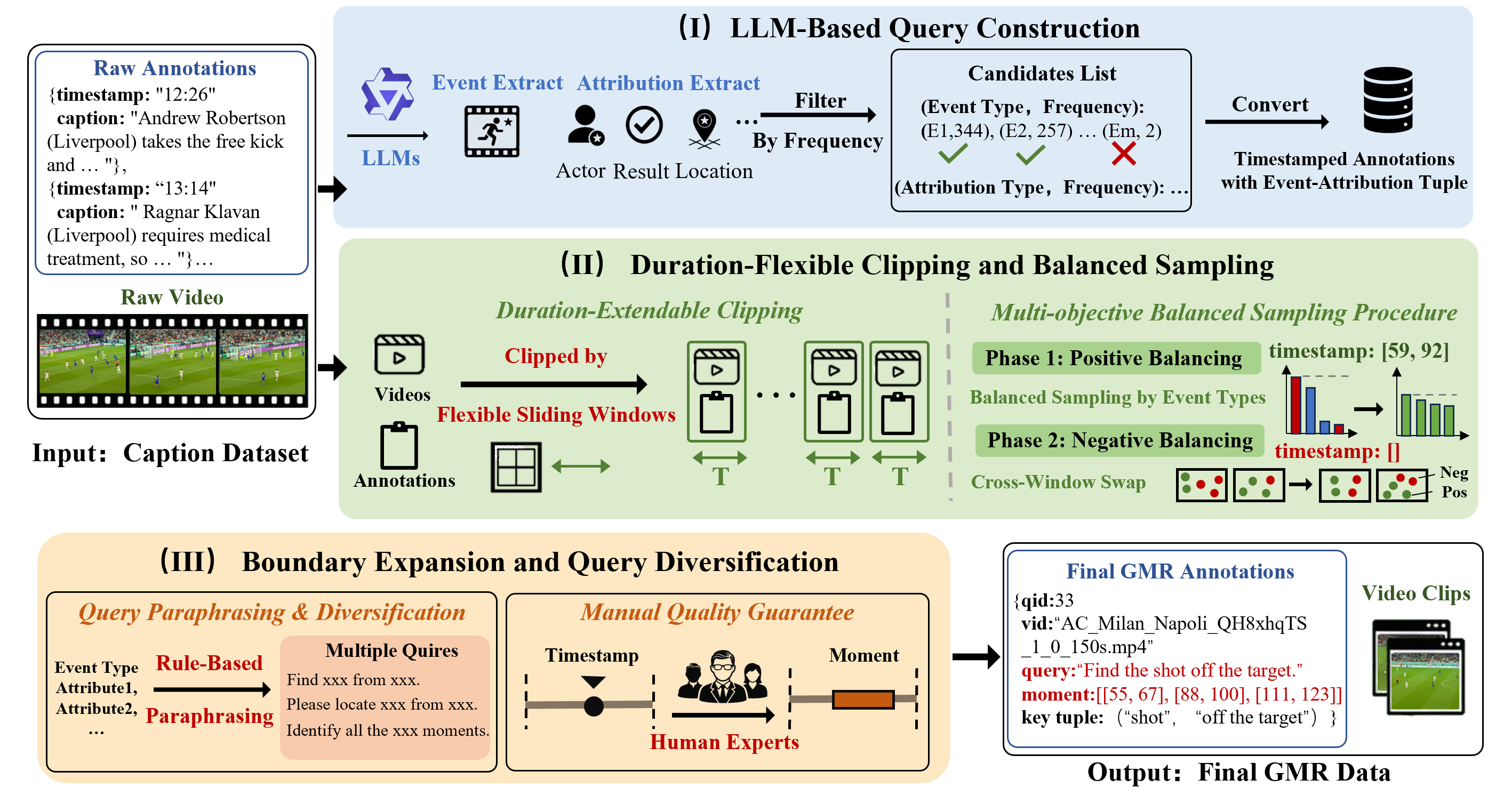}
  \caption{\textbf{Duration-flexible semi-automated pipeline for GMR data construction.} Stage I applies LLMs to extract structured queries from raw timestamp and caption annotations. Stage II segments videos with user-specified sliding-window duration, allowing the same base annotations to produce samples of varying lengths, and applies balanced sampling. Stage III expands point-level timestamps into segment boundaries and diversifies query expressions, followed by expert verification.}
  \label{fig:soccergmr_factory}
\end{figure*}

\section{Benchmark}

To enable the systematic evaluation of GMR, we introduce Soccer-GMR, which covers all three retrieval scenarios: null-set, single-moment, and multi-moment retrieval, together with a unified evaluation protocol. We first formalize the task in Section~\ref{sec:task}, then present the Soccer-GMR dataset in Section~\ref{sec:dataset}, and finally describe the evaluation protocol in Section~\ref{sec:metrics}.

\subsection{Task Definition}
\label{sec:task}

Given a video $V$ and a natural language query $Q$, the goal of GMR is to predict the complete set of temporal segments $\mathcal{T} = \{(t_s^{(i)}, t_e^{(i)})\}_{i=1}^{n}$ in $V$ that correspond to $Q$, where $t_s^{(i)}$ and $t_e^{(i)}$ denote the start and end times of the $i$-th segment. The number of relevant segments $n$ varies across queries:

\begin{itemize}[leftmargin=*]
    \item \textbf{Null-Set Rejection} ($n=0$): No moment is relevant to $Q$, so the model should return an empty set.
    \item \textbf{Single-Moment Retrieval} ($n=1$): Exactly one moment is relevant to $Q$, reducing it to the conventional VMR setting.
    \item \textbf{Multi-Moment Retrieval} ($n>1$): Multiple disjoint moments are relevant to $Q$, and the model should retrieve all of them.
\end{itemize}

Compared with conventional VMR, GMR introduces two additional challenges. \textbf{Null-Set Rejection}: the model is required to correctly reject null-set queries when no moment in the video corresponds to the query, even when such queries share high semantic overlap with positive ones (e.g., "a shot by France" vs.\ "a missed shot by France"), demanding fine-grained compositional reasoning. \textbf{Multi-Moment Retrieval}: the model needs to adaptively determine how many moments to retrieve and maintain sufficient temporal discriminability to identify all distinct occurrences rather than collapsing onto a single dominant moment.

\subsection{Soccer-GMR Dataset}
\label{sec:dataset}
\textbf{Why Soccer?}
We instantiate our GMR benchmark on soccer broadcast footage. Soccer naturally exhibits  all three GMR scenarios: recurring actions yield multi-moment ground truth, while semantically similar but absent events (e.g., a saved shot vs. a deflected shot) produce realistic in-domain negatives, which are more challenging than cross-domain negatives in prior work~\cite{fang2024not,flanagan2025moment}. Its visual complexity (fast motion, fine-grained action distinctions)~\cite{deliege2021soccernet,rao2025towards} further compounds these challenges, while its potential applications to tactical analysis and player assessment provide practical motivation.

\subsubsection{Data Sources and Video Preprocessing.}
We draw data from three sources. \textit{StatsBomb Open Data}~\cite{statsbomb_opendata} and \textit{SoccerReplay-1988}~\cite{rao2025towards} provide timestamp-spot annotations (event-level text with timestamps) and form the primary input to our pipeline. \textit{Sportsmoments}~\cite{kumar2025aligning} provides clip-level caption annotations. We verified its annotation quality on 100 randomly sampled clips with two independent annotators (mean boundary deviation \(<2\)\,s), confirming its compatibility with our benchmark standard.

To standardize input duration and avoid hard-cutting dense events at clip boundaries, all raw footage is segmented into 150-second clips with a 10-second overlap between adjacent clips.

\subsubsection{Data Construction Pipeline.}
Constructing GMR annotations from scratch requires writing queries, finding all relevant moments in each video, and verifying absence, which scales poorly with long videos and dense event distributions. We reduce this cost by leveraging videos with timestamped captions. Such data provides a natural scaffold: timestamps indicate \emph{when} and captions indicate \emph{what}, jointly enabling the scalable construction of structured queries, positive and null-set samples, and segment-level annotations. 

We propose a duration-flexible semi-automated pipeline for GMR data construction, comprising three stages (Figure~\ref{fig:soccergmr_factory}):
\begin{figure}[h]
  \centering
  \includegraphics[width=0.99\columnwidth]{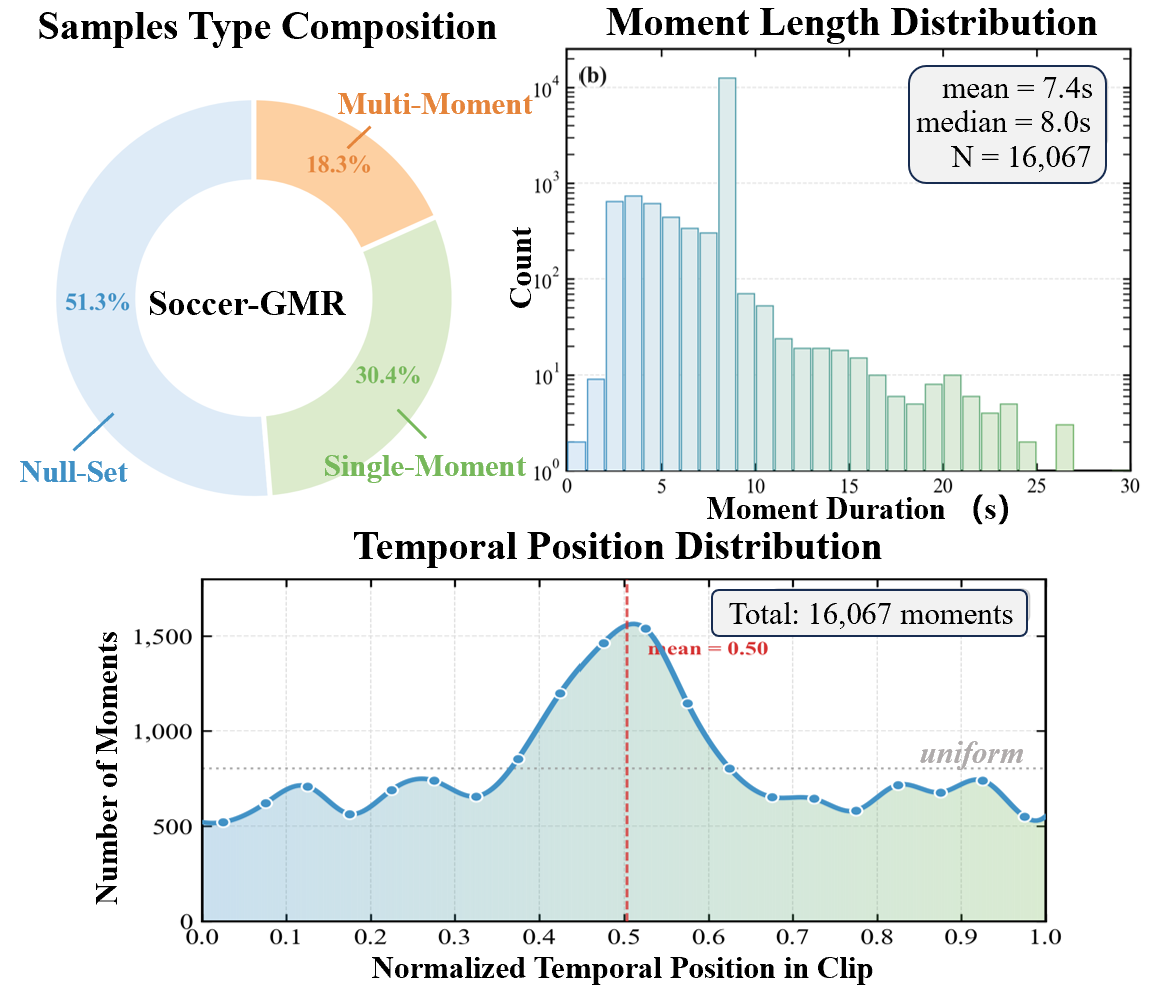}
  \caption{\textbf{Statistics of Soccer-GMR.} Query types include null-set, single-moment, and multi-moment samples (top-left). Ground-truth moments are predominantly short in duration (top-right). Their normalized temporal positions span the entire clip, while showing a noticeable bias toward the middle of the timeline (bottom).}
  \label{fig:dataset_stats}
\end{figure}

\textbf{{Stage I: LLM-Based Query Construction.}} An LLM extracts high-frequency event types and attributes (e.g., actor, result, location) from raw captions, and composes query candidates in the form \(\langle \text{event}, \text{attr}_1, \ldots, \text{attr}_k \rangle\) (\(k \ge 0\)). Candidates are filtered by frequency and utility to form a query vocabulary, then converted into fixed-template base queries with source timestamps and metadata. 

\textbf{{Stage II: Duration-Flexible Clipping and Balanced Sampling.}}
Videos are segmented by a sliding window whose size is freely configurable, so the same base annotations can produce samples at different clip durations without re-annotation (the duration-flexible property). Each clip inherits the Stage-I annotations, with timestamps inside the window treated as positives and those outside as null-set samples. 
Raw segmentation introduces imbalance in three aspects: single-moment queries vastly outnumber multi-moment ones, null-set samples dominate positives, and event-type frequencies follow a long-tail distribution. We apply a two-phase multi-objective balanced sampling procedure (full algorithm in the Appendix) to address all three.

In \emph{Phase~1} (positive balancing), all multi-moment positives are retained while single-moment positives are subsampled at ratio~$\alpha$ relative to the multi-moment count. The budget is allocated across event types by iteratively assigning to the least-represented type until its per-type capacity is reached, so that rare types are preferentially saturated before frequent ones, mitigating long-tail imbalance.

In \emph{Phase~2} (negative balancing), null-set samples are drawn at ratio~$\beta$ relative to positives, allocated proportionally by event type so that the negative subset mirrors the event-type distribution of the retained positives. A cross-window swap then iteratively transfers negatives from surplus windows ($\text{neg}/\text{pos}>\beta$) to deficit windows ($\text{neg}/\text{pos}<\beta$), subject to the invariant that each event type's global negative count is preserved. This prevents individual clips from being dominated by either positives or null-set samples, which could cause the model to overfit to window-specific positive-to-negative priors. 

In this benchmark, we set window length to 150 seconds, motivated by temporal-context limits of current DETR-based SOTA VMR models, and use 10-second overlap to avoid truncating ongoing events. Adjacent clips can be merged to build longer-horizon inputs for future long-video GMR research.

 \begin{table*}[t]
    \centering
    \caption{Comparison of Soccer-GMR with existing VMR benchmarks. $\dagger$\,Statistics are reported from the original paper since the dataset is currently unavailable. *\,Null-set samples are synthetically generated by pairing queries with unrelated videos or by randomly modifying key entities. $\star$\,Duration-flexible: scalable up to 2700\,s (full half-match, 45\,min) by merging adjacent clips.}
    \label{tab:dataset_comparison}  
    \resizebox{0.88\textwidth}{!}{%
    \begin{tabular}{l c c c c c c c c}
    \toprule
    \textbf{Dataset} & \textbf{Domain} & \textbf{\# Queries} &
    \textbf{\begin{tabular}[c]{@{}c@{}}\# Moments /\\ \# Videos\end{tabular}} &
    \textbf{\begin{tabular}[c]{@{}c@{}}Avg. Moment /\\ Query w/ Target\end{tabular}} &
    \textbf{\begin{tabular}[c]{@{}c@{}}Avg. Video\\ Dur.\end{tabular}} &
    \textbf{\begin{tabular}[c]{@{}c@{}}Multi-\\ Moment\end{tabular}} &
    \textbf{\begin{tabular}[c]{@{}c@{}}Null-\\ Set\end{tabular}} &
    \textbf{\begin{tabular}[c]{@{}c@{}}Duration\\ Flexible\end{tabular}} \\
    \midrule
    Charades-STA~\cite{gao2017tall}     & Activity   & 27,847   & 16.1K / 6.7K  & 1.0 & ${\sim}$30s   & $\times$     & $\times$     & $\times$ \\
    DiDeMo~\cite{anne2017localizing}           & Open       & 40,543   & 41.2K / 10.6K  & 1.0 & ${\sim}$27s   & $\times$     & $\times$     & $\times$ \\
    TACoS~\cite{regneri2013grounding}            & Cooking    & 18,818   & 18.8K / 127  & 1.0 & ${\sim}$287s  & $\times$     & $\times$     & $\times$ \\
    ANet-Captions~\cite{krishna2017dense}    & Activity   & 71,953   & 72K / 15K    & 1.0 & ${\sim}$117s  & $\times$     & $\times$     & $\times$ \\
    TVR~\cite{lei2020tvr}              & TV Shows   & 109,480  & 109K / 21.8K  & 1.0 & ${\sim}$76s   & $\times$     & $\times$     & $\times$ \\
    NExT-VMR$^\dagger$~\cite{qin2025generalized} &
    {Open} & {153,191} &
    {229.5K / 9K}  &
    {1.8} & \color{gray}{-} &
    {$\checkmark$} & {$\checkmark$*} & $\times$ \\
    QVHighlights~\cite{lei2021detecting}     & Vlog/News  & 18,367   & 18.5K / 10.2K  & 1.8 & ${\sim}$150s  & $\times$     & $\times$     & $\times$ \\
    QV-M$^2$~\cite{cao2025one}         & Vlog/News  & 2,212    & 6.4K / 1.3K    & 2.9 & ${\sim}$150s  & $\checkmark$ & $\times$     & $\times$ \\
    \midrule
    \textbf{Soccer-GMR (Ours)} & \textbf{Soccer} &
    \textbf{22,119} & \textbf{16.1K / 5.5K} &
    \textbf{1.5} & \textbf{150s$^\star$} &
    \textbf{$\checkmark$} & \textbf{$\checkmark$} & \textbf{$\checkmark$} \\
    \bottomrule
    \end{tabular}%
    }
  \end{table*}

\textbf{{Stage III: Boundary Expansion and Query Diversification.}}
Point-level timestamps from Stage I are expanded into segment-level labels by extending boundaries to fully cover each described event. In the generic pipeline, annotators watch each clip and label start/end boundaries per moment. In our soccer instantiation, instead of full per-moment labeling, we exploit stable duration patterns for same-type soccer events. Annotators first correct timestamps under a unified standard, then estimate event-specific pre/post offsets from a sampled subset and apply them uniformly through rule-based adaptive extension across 29 event types. To validate this strategy, three independent experts annotated the same 300 clips, showing that per-event mean extensions align closely with our rule-derived offsets (full statistics in the appendix).

Fixed template-based queries are further diversified by rule-based paraphrasing into multiple surface forms, improving linguistic diversity and robustness to phrasing variation.

\subsubsection{Data Analysis.}

Soccer-GMR comprises 139 matches, 5.5K video clips, and 22,119 query-moment pairs with 16.1K annotated temporal windows. We use a fixed benchmark split for all experiments, as detailed in the Appendix. Table~\ref{tab:dataset_comparison} compares Soccer-GMR with existing VMR benchmarks. While prior datasets typically assume a single moment per query or rely on synthetic negatives, Soccer-GMR covers all three retrieval scenarios with naturally occurring in-domain negatives. Additionally, its duration-flexible design decouples annotations from clip length, allowing re-segmentation at different durations (e.g., 150\,s to 15\,min) without re-annotation.

As shown in Figure~\ref{fig:dataset_stats}, null-set and positive queries are approximately balanced at a 1:1 ratio (51.3\% vs.\ 48.7\%), with positive queries further split between single-moment and multi-moment cases at roughly 2:1 (30.4\% vs.\ 18.3\%). Ground-truth moments cover a broad range of durations and are distributed across the entire clip timeline, providing diverse temporal coverage.

\subsection{Evaluation Metrics}
\label{sec:metrics}
We organize our metrics into three complementary groups to systematically evaluate GMR: (1) \textbf{Null-Set Rejection}, measuring the ability to correctly reject unanswerable queries. (2) \textbf{Temporal Localization}, assessing temporal grounding accuracy on positive queries. (3) \textbf{Overall GMR Performance}, jointly evaluating both capabilities in a single score. Let $\mathcal{Q}$ denote the full query set, $\mathcal{Q}^{+} = \{q \in \mathcal{Q} \mid |\mathcal{G}(q)| > 0\}$ the positive subset, and $\mathcal{G}(q)$ the set of ground-truth moments for query $q$. Each model produces an existence score $s(q)$: the predicted existence probability for our method, or the maximum predicted window confidence otherwise.

\subsubsection{Null-Set Rejection.}
Since the standard F1 score targets the positive class (correctly retrieving moments) and is not tailored to assess rejection quality, we introduce \textbf{Rej-F1}, which treats the null-set class as the target instead, providing a more direct and intuitive measure of the model's ability to correctly abstain when no relevant moment exists. At operating threshold $\tau$, a query $q$ is classified as null-set if $s(q) \leq \tau$. Rej-F1 is defined as:
\begin{equation}
  \text{Rej-F1} = \frac{2\,\mathrm{TP}_r}{2\,\mathrm{TP}_r + \mathrm{FP}_r + \mathrm{FN}_r},
\end{equation}
where $\mathrm{TP}_r$ counts correctly rejected null-set queries, $\mathrm{FP}_r$ counts positive queries incorrectly rejected, and $\mathrm{FN}_r$ counts null-set queries that the model fails to reject. 

We additionally report \textbf{AUROC}~\cite{fawcett2006introduction} as a threshold-independent measure of the model's ability to discriminate between positive and null-set queries, enabling a fair comparison across models without committing to a specific operating point.

\subsubsection{Temporal Localization.}
Following standard VMR evaluation practice~\cite{lei2021detecting,moon2023query}, we assess localization exclusively on positive queries $\mathcal{Q}^{+}$, so that localization scores purely reflect temporal grounding ability, unaffected by differences in rejection characteristics across models. We adopt established VMR metrics and extend them for multi-moment scenarios.

Let $\mathcal{M}_k(q;\theta)$ denote the set of ground-truth moments matched by the top-$k$ predictions for query $q$ via greedy one-to-one matching at an IoU threshold $\theta$, and let $\mathcal{I} = \{0.50, 0.55, \ldots, 0.95\}$.

\textbf{mR@$k$.} We adopt mean Recall at $k$ (mR@$k$), which generalizes R@$k$~\cite{lei2021detecting} to queries with multiple ground-truth moments:
\begin{equation}
  \text{mR@}k = \frac{1}{|\mathcal{I}|} \sum_{\theta \in \mathcal{I}} \frac{1}{|\mathcal{Q}^{+}|} \sum_{q \in \mathcal{Q}^{+}} \frac{|\mathcal{M}_k(q;\theta)|}{|\mathcal{G}(q)|}.
\end{equation}
For single-moment queries, mR@$k$ reduces to the standard R@$k$ averaged over IoU thresholds. 

\textbf{mR+@$k$.} We observe that what fundamentally distinguishes multi-moment from single-moment retrieval is the ability to retrieve correct segments beyond the first hit. To provide a dedicated measure of this capability, we propose Incremental Recall (mR+@$k$), defined on multi-moment queries $\mathcal{Q}^{m} = \{q \in \mathcal{Q}^{+} \mid |\mathcal{G}(q)| \geq 2\}$:
\begin{equation}
  \text{mR+@}k = \frac{1}{|\mathcal{I}|} \sum_{\theta \in \mathcal{I}} \frac{1}{|\mathcal{Q}^{m}|} \sum_{q \in \mathcal{Q}^{m}} \frac{\max\bigl(0,\,|\mathcal{M}_k(q;\theta)| - 1\bigr)}{|\mathcal{G}(q)| - 1}.
\end{equation}
By excluding the first matched moment from both the numerator and denominator, mR+@$k$ measures the retrieval of additional relevant moments, targeting multi-moment retrieval capability.

\textbf{mAP.} We adopt mean Average Precision following the standard detection protocol~\cite{lei2021detecting}, computed at IoU thresholds $\mathcal{I}$.

\subsubsection{Overall GMR Performance.}
To jointly evaluate rejection and localization, we propose \textbf{G-mIoU@$k$} (Generalized mean IoU at $k$), which assesses end-to-end performance over all queries $\mathcal{Q}$. Using the same operating threshold $\tau$, the model's top-$k$ predictions $\hat{\mathcal{P}}_k(q)$ are gated to $\emptyset$ if $s(q) \leq \tau$. The per-query score is:
\begin{equation}
  \text{IoU}_G(q) {=} \begin{cases}
    1, & \hat{\mathcal{P}}_k(q) {=} \emptyset \;\wedge\; \mathcal{G}(q) {=} \emptyset \\
    \dfrac{\sum_{(i,j)\in \mathcal{M}} \operatorname{IoU}(\hat{p}_i,\, g_j)}{|\hat{\mathcal{P}}_k| + |\mathcal{G}| - |\mathcal{M}|}, & \hat{\mathcal{P}}_k(q) {\neq} \emptyset \;\wedge\; \mathcal{G}(q) {\neq} \emptyset \\
    0, & \text{otherwise}
  \end{cases}
\end{equation}
where $\mathcal{M}$ denotes the greedy one-to-one matching between $\hat{\mathcal{P}}_k(q)$ and $\mathcal{G}(q)$.
\begin{equation}
  \text{G-mIoU@}k = \frac{1}{|\mathcal{Q}|} \sum_{q \in \mathcal{Q}} \text{IoU}_G(q).
\end{equation}
G-mIoU@$k$ assigns a score of 1 for correct rejection, 0 for misclassification between positive and null-set queries, and a set-level IoU between the top-$k$ predictions and all ground-truth moments for correctly accepted positive queries. The set-level IoU penalizes both unmatched predictions and missed ground-truth moments through the union-based denominator, making it particularly suitable for multi-moment evaluation. G-mIoU@$k$ thus serves as a unified measure of overall GMR capability.

In summary, our evaluation framework introduces three targeted metrics (Rej-F1, mR+@$k$, and G-mIoU@$k$) alongside established measures (AUROC, mR@$k$, mAP), extending conventional VMR evaluation to cover null-set rejection, multi-moment localization, and end-to-end GMR performance.

\section{Method}
We consider two modeling approaches for GMR: a lightweight adapter for classical VMR methods (Section~\ref{sec:GMR Adapter}) and RL-based fine-tuning for generative MLLMs (Section~\ref{sec:MLLM Fine-Tuning via GRPO}).
  
\subsection{GMR Adapter}

\label{sec:GMR Adapter}
\begin{figure}[t]
\centering
\includegraphics[width=\linewidth]{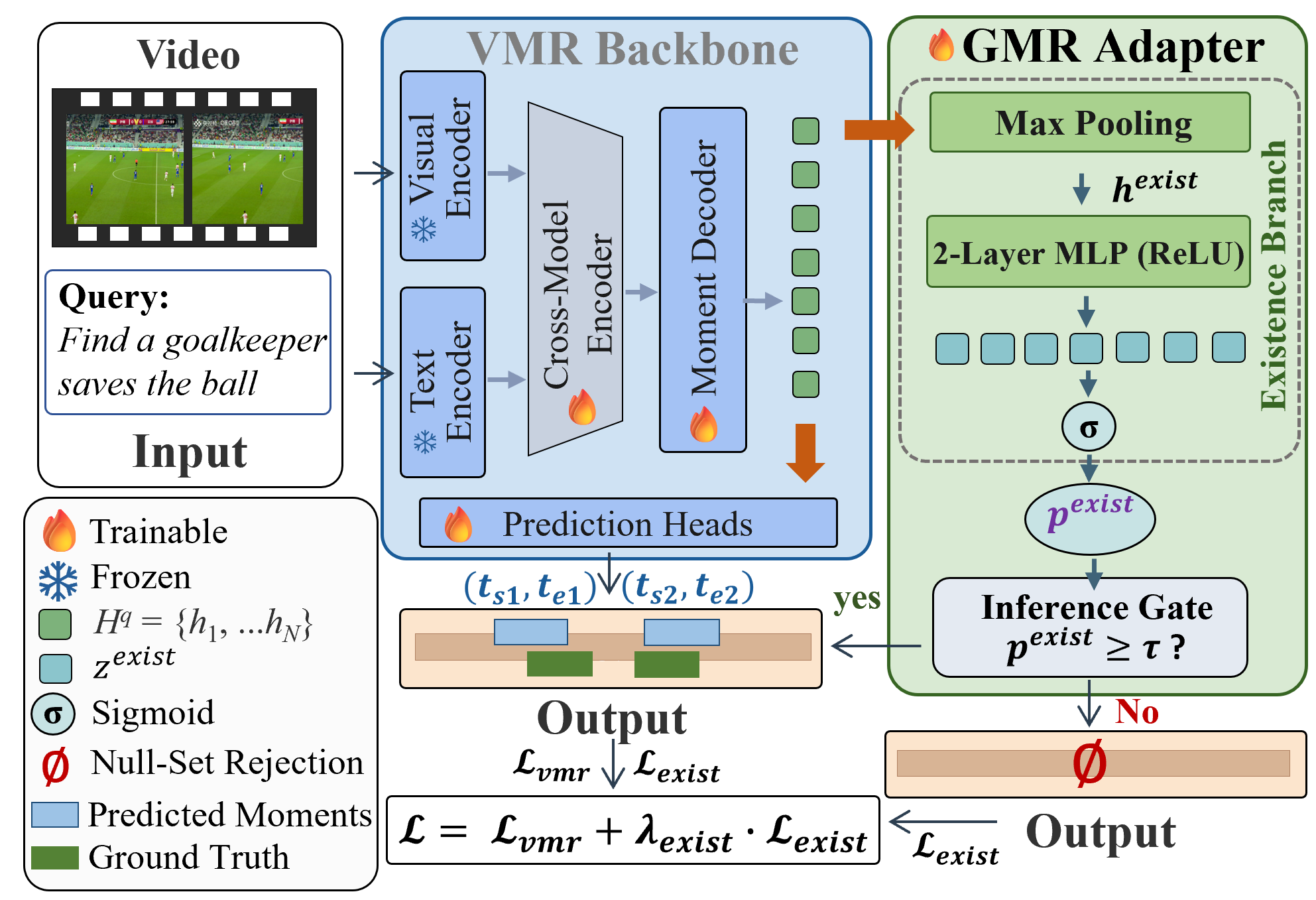}
\caption{\textbf{Architecture of the GMR Adapter.} A parallel existence branch computes $p^{\text{exist}}$ from cross-modal representations $H^q$ via max-pooling and a two-layer MLP. At inference, $p^{\text{exist}}$ is compared against a threshold $\tau$ to gate the backbone's moment predictions, enabling null-set rejection without modifying the original architecture.}
\label{fig:m}
\end{figure}
  
\textbf{Overview.} Discriminative VMR models share a common moment decoding stage that produces query-conditioned cross-modal representations, reflecting the model's response to the query after attending to the full video. We observe that these representations provide a natural anchor for existence estimation: strong slot activations indicate relevant content, while uniformly weak activations indicate a null-set query. Building on this, we propose the GMR Adapter, a lightweight plug-and-play module that attaches a parallel existence branch alongside the prediction heads of VMR backbones without modifying the backbone architecture (Figure~\ref{fig:m}), and is compatible with backbones that expose such representations after the moment decoding stage.

\subsubsection{Existence Branch.}
Let $H^q = {h_1, \dots, h_N},\ h_i \in \mathbb{R}^d$ denote the query embeddings from the last decoder layer, where $N$ is the number of query
slots and $d$ is the hidden dimension. To obtain a single query-video-level representation, we apply max-pooling over the query dimension:
\begin{equation}
 h^{\text{exist}} = \max_{i=1,\dots,N}, h_i \in \mathbb{R}^d.
\end{equation}
Max-pooling selects the strongest slot response across all $N$ candidates, which serves as a natural indicator of existence: a strongly activated
slot signals a relevant moment, while uniformly weak activations indicate a null-set query.
 
The pooled representation is passed through a two-layer MLP with ReLU activation to produce a scalar existence logit $z^{\text{exist}}$, from
which the existence probability is obtained via sigmoid:
\begin{equation}
  z^{\text{exist}} = \mathrm{MLP}(h^{\text{exist}}), \qquad p^{\text{exist}} = \sigma(z^{\text{exist}}) \in (0, 1),
\end{equation} 
where $p^{\text{exist}}$ estimates the probability that at least one relevant moment exists for the current query-video pair. The existence branch runs in parallel with the backbone's original localization and classification heads, sharing $H^q$ as input without modifying the backbone's forward computation. For backbones without explicit decoder query slots (e.g., FlashVTG), $H^q$ is derived from the model's equivalent cross-modal representation.

\subsubsection{Training Objective.} For each training sample, we construct a binary existence label from its ground-truth moment set $\mathcal{G}$:
 \begin{equation}
     y^{\text{exist}} = \begin{cases} 1, & |\mathcal{G}| > 0 \\ 0, & |\mathcal{G}| = 0 \end{cases}.
 \end{equation}
Null-set samples are included in the same batch alongside positive samples without any modification to the backbone's training procedure: Hungarian matching produces an empty assignment for null-set samples, so $\mathcal{L}_{\text{vmr}}$ contributes no gradient for these samples, while positive samples jointly optimize both $\mathcal{L}_{\text{vmr}}$ and $\mathcal{L}_{\text{exist}}$, and null-set samples receive supervision from $\mathcal{L}_{\text{exist}}$ alone. The overall loss is:
\begin{equation}
      \mathcal{L} = \mathcal{L}_{\text{vmr}} + \lambda_{\text{exist}} \cdot \mathcal{L}_{\text{exist}},
\end{equation}

\begin{table*}[t]
\centering
\tiny
\renewcommand{\arraystretch}{0.8}
\caption{Main results on the Soccer-GMR benchmark test set ($\tau{=}0.4$). G-mIoU@$k$ evaluates end-to-end GMR ability on all samples. AUROC measures threshold-free rejection ability, and Rej-F1 reports rejection quality at the main operating point. mAP, mR@1, mR@5, and mR+@5 evaluate positive-query temporal localization and multi-moment retrieval.}
\label{tab:main_results_final}
\resizebox{0.85\linewidth}{!}{%
\begin{tabular}{l cc cccc cc}
\toprule
& \multicolumn{2}{c}{\textit{Null-Set Rejection}} & \multicolumn{4}{c}{\textit{Temporal Localization}} & \multicolumn{2}{c}{\textit{Overall GMR}} \\
\cmidrule(lr){2-3} \cmidrule(lr){4-7} \cmidrule(lr){8-9}
\textbf{Model} & AUROC & Rej-F1 & mAP & mR@1 & mR@5 & mR+@5 & G-mIoU@1 & G-mIoU@3 \\
\midrule
Moment-DETR~\cite{lei2021detecting} & 69.92 & 0.00 & 6.98  & 3.78  & 10.92 & 0.78 & 5.39 & 2.47 \\
QD-DETR~\cite{moon2023query}        & 64.71 & 0.00 & 9.62  & 6.30  & 12.81 & 4.47 & 8.48 & 4.22 \\
CG-DETR~\cite{moon2023correlation}  & 72.27 & 0.00 & 15.85 & 10.81 & 21.22 & 9.03 & 11.83 & 6.00 \\
EATR~\cite{jang2023knowing}         & 70.99 & 0.80 & 18.48 & 11.69 & 25.27 & 11.81 & 12.94 & 6.67 \\
FlashVTG~\cite{cao2025flashvtg}     & 57.33 & 7.12 & \underline{23.61} & \underline{14.56} & \underline{33.06} & \underline{15.30} & 15.41 & 8.21 \\
\midrule
Moment-DETR-GMR (Ours) & 72.09 & \textbf{64.01} & 7.52  & 3.51  & 12.96 & 0.84 & 35.84 & \underline{32.89} \\
EATR-GMR (Ours)        & \textbf{79.11} & \underline{62.10} & 18.56 & 12.78 & 24.43 & 13.97 & \underline{37.89} & 31.95 \\
FlashVTG-GMR (Ours)    & \underline{74.00} & 61.72 & \textbf{24.62} & \textbf{15.08} & \textbf{33.36} & \textbf{19.10} & \textbf{39.58} & \textbf{33.53} \\
\bottomrule
\end{tabular}}
\end{table*}

\noindent where $\mathcal{L}_{\text{vmr}}$ is the backbone's original VMR training objective and $\lambda_{\text{exist}}$ is a scalar weight. The existence branch is supervised via binary cross-entropy:
 \begin{equation}
     \mathcal{L}_{\text{exist}} = \mathrm{BCEWithLogits}(z^{\text{exist}}, y^{\text{exist}}).
 \end{equation}

The adapter requires only that the backbone exposes cross-modal representations after the moment decoding stage and that $\mathcal{L}_{\text{vmr}}$ supports an additive auxiliary term. Since $\mathcal{L}_{\text{exist}}$ attaches as an independent additive term without interacting with any backbone-specific loss component, these conditions are satisfied by all three backbones we evaluate.

\subsubsection{Inference.}
At inference, the model produces an existence score $p^{\text{exist}}$ alongside the backbone's span predictions, with a threshold $\tau$ gating the final output:
 \begin{equation}
       \hat{\mathcal{T}} = \begin{cases} 
\emptyset, & p^{\text{exist}} < \tau \\ 
\{ \hat{t}_s^{(i)}, \hat{t}_e^{(i)} \}_{i=1}^{N}, & p^{\text{exist}} \geq \tau
\end{cases}.
 \end{equation}

When $p^{\text{exist}} \geq \tau$, the backbone's original prediction pipeline is used unchanged, naturally supporting both single-moment and multi-moment retrieval without additional post-processing.

\subsection{GRPO with a GMR-Tailored Reward}
\label{sec:MLLM Fine-Tuning via GRPO}
To adapt generative MLLMs to the structured prediction requirements of GMR, 
we design a GMR-tailored GRPO reward. Specifically, we leverage a task-specific rule-based reward within GRPO and use it to fine-tune the MLLMs with LoRA~\cite{hu2022lora}, directly capturing retrieval, localization, and rejection behavior. 

Concretely, for non-empty targets, the reward combines two metric-aligned terms: a retrieval term based on $\mathrm{mR@}k$ and a localization term based on $\mathrm{mIoU@}k$, with $k \in \{1,2,3\}$. For each $k$, predicted windows are greedily matched to unmatched ground-truth windows, and performance is aggregated across multiple IoU thresholds. This design encourages the model not only to retrieve the correct number of moments, but also to localize them precisely.

We further incorporate explicit handling of null-set cases. When the ground truth contains no relevant moment, correctly predicting an empty set receives a positive reward, whereas false positives are penalized. Conversely, if relevant moments exist but the model predicts no window, the sample receives a negative reward. This gives GRPO direct supervision for rejection behavior, which is central to GMR but absent from standard VMR-style training.

Finally, we apply validity penalties to suppress degenerate outputs, including excessive predictions, out-of-range boundaries, and zero-length spans, and malformed outputs receive a failure penalty. Overall, the reward encourages three properties simultaneously: correct rejection on null-set queries, high recall over multiple relevant moments, and precise temporal localization. Training details are provided in Section~\ref{sec:setup}.

\section{Experiments}
\subsection{Experimental Setup}
\label{sec:setup}
\textbf{Baselines.} We compare against five state-of-the-art VTG models, Moment-DETR~\cite{lei2021detecting}, QD-DETR~\cite{moon2023query}, CG-DETR~\cite{moon2023correlation}, EaTR~\cite{jang2023knowing}, and FlashVTG~\cite{cao2025flashvtg}, and further evaluate GMR-extended variants (Moment-DETR-GMR, EaTR-GMR, and FlashVTG-GMR), which augment the respective base models with the GMR Adapter (Section~\ref{sec:GMR Adapter}) for explicit null-set rejection. All discriminative models are trained on the training split. For the MLLM paradigm, we evaluate Qwen3-VL (4B, 8B, 32B)~\cite{bai2025qwen3} in the zero-shot setting, alongside temporal grounding specialist model TRACE~\cite{guo2024trace} and video temporal understanding model TimeChat~\cite{ren2024timechat}, and additionally fine-tune Qwen3-VL-4B with GRPO (Section~\ref{sec:MLLM Fine-Tuning via GRPO}).

\textbf{Implementation Details.} All models process video frames sampled at 1~fps. For discriminative models, frames are encoded with CLIP~\cite{radford2021learning} and SlowFast~\cite{feichtenhofer2019slowfast} features, and queries with the CLIP text encoder. For fair comparison, all discriminative baselines and their GMR variants share these input representations and are trained with a learning rate of $3\times10^{-5}$. For the GMR Adapter, we set the existence-loss coefficient to $\lambda=1.0$ and select the inference threshold $\tau=0.4$ based on validation performance.
For MLLMs, all Qwen3-VL variants use thinking mode for inference. For GRPO, we fine-tune Qwen3-VL-4B-Instruct with LoRA and set the maximum generation length to 1024 tokens. GRPO training is conducted on three A800 80\,GB GPUs.

\begin{table*}[ht]
    \centering
    \caption{Query style robustness results. All reformulations preserve core semantic content (event type and attribute constraints), and only surface form and length vary. \textbf{Bold}: best per model per
    metric. {($-\Delta$)}: drop relative to original.}
    \label{tab:prompt}
    \small
    \setlength{\tabcolsep}{4pt}
    \begin{tabular}{lc ccc ccc ccc}
    \toprule
    & &
    \multicolumn{3}{c}{\textbf{FlashVTG-GMR}} &
    \multicolumn{3}{c}{\textbf{EaTR-GMR}} &
    \multicolumn{3}{c}{\textbf{Moment-DETR-GMR}} \\
    \cmidrule(lr){3-5}\cmidrule(lr){6-8}\cmidrule(lr){9-11}
    \textbf{Style} & \textbf{Avg.\ Len.}
      & AUROC & mAP & mR+@5
      & AUROC & mAP & mR+@5
      & AUROC & mAP & mR+@5 \\
    \midrule
    Original       & ${\sim}7$
      & \textbf{74.00} & \textbf{24.62} & \textbf{19.10}
      & \textbf{79.11} & 18.56          & \textbf{13.97}
      & \textbf{72.09} & 7.52           & \textbf{0.84} \\
    B: Question    & ${\sim}9$
      & 71.63 & 23.20 & 15.88
      & 76.78 & \textbf{18.74} & 10.82
      & 71.07 & \textbf{8.01} & 0.45 \\
    C: Noun Phrase & ${\sim}8$
      & 69.59 & 23.32 & 16.56
      & 77.63 & 17.07 & 12.29
      & 71.19 & 7.17 & 0.51 \\
    \midrule
    D: Keyword     & ${\sim}3$
      & 54.64{\scriptsize\,{($-$19.36)}}
      & 17.85{\scriptsize\,{($-$6.77)}}
      & 8.47{\scriptsize\,{($-$10.63)}}
      & 68.41{\scriptsize\,{($-$10.70)}}
      & 17.46{\scriptsize\,{($-$1.10)}}
      & 11.08{\scriptsize\,{($-$2.89)}}
      & 69.57{\scriptsize\,{($-$2.52)}}
      & 6.47{\scriptsize\,{($-$1.05)}}
      & 0.70{\scriptsize\,{($-$0.14)}} \\
    E: Verbose     & ${\sim}28$
      & 69.98{\scriptsize\,{($-$4.02)}}
      & 17.87{\scriptsize\,{($-$6.75)}}
      & 13.24{\scriptsize\,{($-$5.86)}}
      & 73.70{\scriptsize\,{($-$5.41)}}
      & 17.54{\scriptsize\,{($-$1.02)}}
      & 11.00{\scriptsize\,{($-$2.97)}}
      & 63.44{\scriptsize\,{($-$8.65)}}
      & 4.42{\scriptsize\,{($-$3.10)}}
      & 0.60{\scriptsize\,{($-$0.24)}} \\
    \bottomrule
    \end{tabular}
    \end{table*}

\vspace{-0.2\baselineskip}
\subsection{Main Results}
\label{sec:main}
Table~\ref{tab:main_results_final} reports results on the benchmark test set. The GMR Adapter consistently improves all three backbones, achieving substantial gains in rejection ability while maintaining or slightly improving localization quality.

\textbf{Rejection ability.}
Without an explicit rejection mechanism, conventional VMR baselines lack supervision for null-set queries, yielding Rej-F1 scores no higher than 7.12. The GMR Adapter improves AUROC by up to 16.67\% (FlashVTG-GMR) and achieves Rej-F1 of 61.72--64.01 across all three backbones, indicating substantially improved discriminative capacity between positive and null-set queries.

\textbf{Temporal localization.}
Beyond rejection, the GMR Adapter preserves localization quality, with temporal localization metrics remaining comparable to or slightly exceeding those of the base models across all three backbones, suggesting that the auxiliary existence objective complements rather than competes with the localization loss.

\textbf{Multi-moment retrieval.}
While the GMR Adapter yields consistent improvements (+3.80\% on FlashVTG mR+@5), the absolute mR+@5 values remain low across all models, with the highest being 19.10, indicating that current architectures still struggle to reliably localize multiple distinct moments for the same query. Multi-moment retrieval thus remains a key open challenge.

\begin{table}[htbp]
    \centering
    \caption{MLLM evaluation on the Soccer-GMR benchmark test set ($\tau{=}0.4$). Top: zero-shot, bottom: fine-tuned via GRPO (Section~\ref{sec:MLLM Fine-Tuning via GRPO}). \textbf{Bold}: best in column.}
    \label{tab:mllm}
    \small
    \setlength{\tabcolsep}{4pt}
    \resizebox{\linewidth}{!}{
    \begin{tabular}{lccccc}
    \toprule
    \textbf{Model} & AUROC & Rej-F1 & mAP & mR@1 & mR+@5 \\
    \midrule
    Qwen3-VL-4B~\cite{bai2025qwen3}
      & 47.66 & 11.54 & 2.15 & 1.65 & 0.12 \\
    Qwen3-VL-8B~\cite{bai2025qwen3}
      & 52.60 & \underline{43.11} & 1.94 & 1.54 & \underline{0.16} \\
    Qwen3-VL-32B~\cite{bai2025qwen3}
      & \textbf{57.75} & \textbf{54.26} & \underline{2.76} & \textbf{2.42} & 0.06 \\
    \midrule
    TRACE~\cite{guo2024trace}\hspace{0.3em}{\footnotesize\textit{\textcolor{gray}{ICLR'25}}} \\
      & 50.00 & 0.00 & 1.40 & 1.39 & 0.00 \\
    TimeChat~\cite{ren2024timechat}\hspace{0.3em}{\footnotesize\textit{\textcolor{gray}{CVPR'24}}} \\
      & 50.85 & 4.72 & 0.49 & 0.39 & 0.00 \\
    \midrule
    Qwen3-VL-4B (GRPO)
      & \underline{53.21} & 13.83 & \textbf{2.91} & \underline{1.97} & \textbf{1.18} \\
    \bottomrule
    \end{tabular}
    }
    \end{table}

\subsection{MLLM Evaluation}
\label{sec:mllm}
As shown in Table~\ref{tab:mllm}, the best-performing MLLM achieves notably lower localization than the best discriminative model FlashVTG-GMR, suggesting that generative MLLMs face substantial challenges in temporal grounding under the GMR setting.

\textbf{Specialist temporal grounding models.}
TRACE and TimeChat underperform Qwen3-VL across nearly all metrics, with both showing near-random rejection ability (AUROC $\leq$ 50.85). These results suggest that methods developed for conventional single-moment VMR do not transfer well to GMR.

\textbf{Effect of model scale.}
Within the Qwen3-VL family, larger models exhibit stronger rejection ability, as reflected by AUROC gains from 47.66 (4B) to 52.60 (8B) and 57.75 (32B). However, localization performance remains at very low absolute levels across all model scales, with mR@1 only improving marginally from 1.65 to 2.42. These results indicate that increasing model scale does not meaningfully resolve the fine-grained temporal grounding challenges posed by GMR.

\textbf{MLLM fine-tuning.}
To investigate whether task-specific fine-tuning can close this gap,
we fine-tune Qwen3-VL-4B with GRPO. GRPO yields consistent gains
across all metrics, with rejection and localization improving
simultaneously rather than trading off. Notably, the 4B
GRPO-fine-tuned model surpasses the $8\times$ larger 32B zero-shot
model on localization and multi-moment retrieval (mAP $2.91$ vs.\
$2.76$, mR+@5 $1.18$ vs.\ $0.06$), suggesting that task-specific
fine-tuning particularly benefits these capabilities, whereas
rejection still scales with model size. However, the localization
gap relative to the best discriminative model remains substantial
(mAP $2.91$ vs.\ $24.62$), suggesting that task-specific RL can narrow but not substantially close the localization gap of generative MLLMs.

\subsection{Query Style Robustness}
\label{sec:prompt}
We evaluate all three GMR models under five query reformulations in two categories: phrasing variants (B/C), which alter sentence structure at comparable lengths, and length variants (D/E), which substantially shorten or lengthen the query. All reformulations
preserve the same core semantic content. Details and examples are provided in the Appendix.

Results are shown in Table~\ref{tab:prompt}. Across phrasing
variants, all models exhibit stable performance, with AUROC varying
by at most 4.41 points and mAP varying by at most 1.49 points. However, length
variants cause consistent degradation across all models (e.g.,
FlashVTG-GMR AUROC drops by 19.36 points under keyword queries, and
mAP decreases by 6.75 points under verbose queries), suggesting that
query length is a more critical factor than phrasing for GMR
robustness.

\section{Conclusion}
In this paper, we present a systematic study of Generalized Moment Retrieval (GMR), extending conventional VMR to handle queries with any number of relevant moments, including none. We introduce Soccer-GMR, a large-scale benchmark with realistic in-domain negatives and multi-moment annotations, accompanied by a semi-automated construction pipeline that reduces annotation costs and a unified evaluation protocol with complementary metrics. We further propose the GMR Adapter for discriminative VMR backbones and a GMR-tailored GRPO reward for MLLM fine-tuning, establishing baselines along both paradigms.


\appendix
\section*{Appendix}

\section{Soccer-GMR Benchmark Construction Details}
\label{sec:benchmark_details}

\subsection{LLM-Based Query Construction}
\label{sec:llm_prompt}

Stage~I of the annotation pipeline (Sec.~3.2.2) employs
Qwen3-8B-Instruct~\cite{bai2025qwen3} to convert unstructured video captions into
structured event-attribute records. The extraction proceeds
through four steps.

\paragraph{Step 1: Event and Attribute Extraction.}
Each input record typically includes a video identifier, a raw
caption, and a point-level timestamp (in seconds):
\begin{verbatim}
  {vid: "AC_Milan_Napoli_QH8xhqTS_1.mp4",
   caption: "Andrew Robertson takes the
             free kick and ...",
   timestamp: 764}
\end{verbatim}
The LLM parses the caption and extracts all identifiable events,
each decomposed into an event type and a set of semantic attributes
(e.g., \textit{actor}, \textit{result}, \textit{location}).
The output is a structured tuple
$\langle \textit{event\_type},\; \textit{attr}_1,\; \dots,\;
\textit{attr}_k \rangle$ ($k \ge 0$).

\paragraph{Step 2: Semantic Unification.}
Different surface realizations of the same event semantics are merged
into canonical forms prior to frequency counting.
For instance, "shoots wide of the post" and "shot goes wide"
are both normalized to \texttt{("shot",\;"off the target")}.
This step ensures that frequency statistics faithfully reflect true
event prevalence rather than lexical variation.

\paragraph{Step 3: Frequency-Based Filtering.}
After unification, event types and attribute values are counted across
the entire corpus. Candidates below a frequency threshold are
discarded, retaining only high-frequency, semantically meaningful
event-attribute combinations. This yields a compact
\emph{query vocabulary} that is both representative and statistically
reliable.

\paragraph{Step 4: Aggregation and Template Conversion.}
Surviving tuples are grouped by \texttt{key\_tuple} per video,
collecting all matching timestamps into a single list:
\begin{verbatim}
  {vid: "AC_Milan_Napoli_QH8xhqTS_1.mp4",
   key_tuple: ("shot", "off the target"),
   timestamp: [59, 92, ..., 746]}
\end{verbatim}
Each \texttt{key\_tuple} is then converted into a fixed-template
natural-language query with its source timestamps and metadata,
which serves as input to Stage~II (duration-flexible clipping) and
Stage~III (query diversification).

\paragraph{Core Prompt Template.}
The extraction prompt is fully domain-agnostic: the LLM infers
relevant event types and attributes directly from caption content,
requiring no manual adaptation across domains.

\begin{tcolorbox}[colback=gray!5, colframe=gray!60,
  fontupper=\small\ttfamily,
  boxsep=2pt, top=2pt, bottom=2pt, left=4pt, right=4pt,
  title={\small\sffamily Extraction Prompt Template}]
\#\# Role\\
You are an expert event extraction assistant,
specialized in identifying structured events and
their semantic attributes from video narration
text.\\[4pt]
\#\# Task\\
Given a narration caption from a video, extract all
described events as structured records. For each event,
output:\\
- event\_type: a concise canonical label for the action\\
- attributes: a dictionary of relevant properties
  (e.g., actor, result, location)\\[4pt]
\#\# Rules\\
1. Use consistent, canonical expressions; if two phrases
   describe the same semantics, use the same label.\\
2. One caption may contain zero, one, or multiple events.\\
3. Output strictly in JSON format.\\[4pt]
\#\# Example\\
Input: "Andrew Robertson takes the free kick and
the shot goes off the target."\\
Output:\\{}
[\{\\
\quad "event\_type": "shot",\\
\quad "attributes": \{\\
\qquad "actor": "Andrew Robertson",\\
\qquad "result": "off the target"\\{}
\quad \}\}]
\end{tcolorbox}

\subsection{Multi-Objective Balanced Sampling}
\label{sec:balanced_sampling}

We provide the complete pseudocode for the two-phase balanced sampling
procedure outlined in Sec.~3.2.2.
Algorithm~\ref{alg:waterfill} describes the capacity-constrained
uniform allocation subroutine used in Phase~1, and
Algorithm~\ref{alg:balanced_sampling} presents the full procedure.

\begin{algorithm*}[t]
\caption{\textsc{WaterFill}: Capacity-Constrained Uniform Allocation}
\label{alg:waterfill}
\begin{algorithmic}[1]
\Require Per-type capacities $\{c_e\}_{e \in \mathcal{E}}$, total budget $B$
\Ensure Allocation $\{a_e\}$ s.t.\ $a_e \le c_e,\;\forall e$ and $\textstyle\sum_e a_e = \min\!\bigl(B,\,\sum_e c_e\bigr)$
\State $a_e \leftarrow 0$ for all $e \in \mathcal{E}$
\While{$B > 0$ \textbf{and} $\exists\, e\!: a_e < c_e$}
    \State $e^{\star} \leftarrow \arg\min_{e:\, a_e < c_e} a_e$ \Comment{least-filled type with remaining capacity}
    \State $a_{e^{\star}} \leftarrow a_{e^{\star}} + 1$;\; $B \leftarrow B - 1$
\EndWhile
\State \Return $\{a_e\}$
\end{algorithmic}
\end{algorithm*}

\begin{algorithm*}[t]
\caption{Multi-Objective Balanced Sampling}
\label{alg:balanced_sampling}
\begin{algorithmic}[1]
\Require Windowed clips $\{W_j\}_{j=1}^{N}$, with every sample $x$ classified as multi-moment positive ($x \!\in\! \mathcal{P}^{\mathrm{m}}$), single-moment positive ($x \!\in\! \mathcal{P}^{\mathrm{s}}$), or null-set sample ($x \!\in\! \mathcal{N}$); single-to-multi ratio $\alpha$; negative-to-positive ratio $\beta$; max rounds $T$; max swaps $S$
\Ensure Balanced dataset $\mathcal{D}$

\Statex
\Statex \hspace{-1.2em}\textbf{Phase\;1:\;Single--multi positive balancing}
\State $\mathcal{P} \leftarrow \mathcal{P}^{\mathrm{m}}$
    \Comment{retain all multi-moment positives}
\State $B \leftarrow \left\lfloor \alpha \cdot |\mathcal{P}^{\mathrm{m}}| \right\rfloor$
    \Comment{global single-moment budget}
\State $c_e \leftarrow |\{x \in \mathcal{P}^{\mathrm{s}} : \mathrm{type}(x) = e\}|$
    for each $e \in \mathcal{E}$
    \Comment{per-type capacity}
\State $\{a_e\} \leftarrow \textsc{WaterFill}\!\bigl(\{c_e\},\, B\bigr)$
    \Comment{Alg.\;\ref{alg:waterfill}}
\For{each event type $e \in \mathcal{E}$}
    \State Sample $a_e$ items uniformly from
           $\{x \in \mathcal{P}^{\mathrm{s}} : \mathrm{type}(x)=e\}$ and
           add to $\mathcal{P}$
\EndFor

\Statex
\Statex \hspace{-1.2em}\textbf{Phase\;2:\;Positive--negative balancing}

\Statex \hspace{-0.4em}\textit{2a.\;Global proportional sampling}
\State $\mathcal{N}_{\!\mathrm{sel}} \leftarrow \emptyset$
\For{each event type $e \in \mathcal{E}$}
    \State $n_e \leftarrow
      \left\lfloor \beta \cdot |\{x \in \mathcal{P} : \mathrm{type}(x) = e\}| \right\rfloor$
      \Comment{type-level negative target}
    \State Sample $\min\!\bigl(n_e,\;|\{x \in \mathcal{N} : \mathrm{type}(x) = e\}|\bigr)$
           negatives of type $e$ and add to $\mathcal{N}_{\!\mathrm{sel}}$
\EndFor
\If{$|\mathcal{N}_{\!\mathrm{sel}}| < \left\lfloor \beta \cdot |\mathcal{P}| \right\rfloor$}
    \State Randomly supplement from $\mathcal{N} \setminus \mathcal{N}_{\!\mathrm{sel}}$
           until
           $|\mathcal{N}_{\!\mathrm{sel}}| \ge \left\lfloor \beta \cdot |\mathcal{P}| \right\rfloor$
           or the pool is exhausted
\EndIf

\Statex
\Statex \hspace{-0.4em}\textit{2b.\;Cross-window swap refinement}
\State $s \leftarrow 0$
    \Comment{swap counter}
\For{round $= 1, \ldots, T$}
    \State $\mathcal{D}^{+} \!\leftarrow\! \bigl\{j : |\mathcal{N}_{\!\mathrm{sel}}^{W_j}|
           > \beta\,|\mathcal{P}^{W_j}|\bigr\}$,\;
           $\mathcal{D}^{-} \!\leftarrow\! \bigl\{j : |\mathcal{N}_{\!\mathrm{sel}}^{W_j}|
           < \beta\,|\mathcal{P}^{W_j}|\bigr\}$
    \If{$\mathcal{D}^{+}\!=\!\emptyset$ \textbf{or}
        $\mathcal{D}^{-}\!=\!\emptyset$}
        \textbf{break}
    \EndIf
    \State Sort $\mathcal{D}^{+}$ by surplus desc.,\;
           $\mathcal{D}^{-}$ by deficit desc.
    \State \textit{progress} $\leftarrow$ \textbf{false}
    \For{each $(d, v) \in \mathcal{D}^{+} \!\times\! \mathcal{D}^{-}$}
        \If{$\exists\, e$\!: window $d$ has a selected neg of type $e$
            \textbf{and} window $v$ has an unselected neg of type $e$}
            \State Deselect one type-$e$ neg from window $d$;\;
                   select one type-$e$ neg into window $v$
            \State $s \leftarrow s+1$;\;
                   \textit{progress} $\leftarrow$ \textbf{true}
                   \Comment{per-type global count invariant}
        \EndIf
        \If{$s \ge S$} \textbf{break} \EndIf
    \EndFor
    \If{$\neg$\textit{progress}} \textbf{break} \EndIf
\EndFor

\Statex
\State \Return $\mathcal{D} \leftarrow \mathcal{P} \cup \mathcal{N}_{\!\mathrm{sel}}$
\end{algorithmic}
\end{algorithm*}

\paragraph{Implementation Note.}
In our soccer instantiation, the matching granularity in Steps~2a and~2b is refined from event type to the \emph{semantic group} $\langle\text{event},\,\text{attribute}\rangle$ (e.g., $\langle\textit{pass},\,\textit{Player\;A}\rangle$), falling back to event-type matching when the finer group has insufficient candidates. This exploits the observation that windows derived from the same source video often share identical semantic groups, improving the effectiveness of cross-window swaps.

\subsection{Boundary Expansion Quality}
\label{sec:boundary_expansion}

To validate the rule-based boundary expansion in Stage~III
(Sec.~3.2.2), three annotators independently labeled
approximately 300 clips. Table~\ref{tab:boundary_exp} compares
the annotators' observed expansion with the parameters adopted
in our pipeline for the major event types.

\paragraph{Observations.}
(1)~The adopted expansion parameters closely align with the
annotators' observed values across all event types, confirming
that the rule-based expansion produces boundaries consistent
with human judgment.
(2)~Fast on-pitch actions (save, dribble, tackle, block,
clearance, shot) exhibit compact and stable expansion
(forward 2--5\,s, backward 3--4\,s) with low cross-annotator
variance, while ceremonial events (yellow card, substitution)
naturally require larger windows yet still show strong
inter-annotator agreement (e.g., yellow card backward
std\,=\,0.4\,s).

\begin{table}[htbp]
\centering
\caption{Boundary expansion parameters vs.\ human annotations.
Fwd/Bwd: forward/backward expansion in seconds.}
\label{tab:boundary_exp}
\small
\resizebox{\columnwidth}{!}{%
\begin{tabular}{l cc cc}
\toprule
 & \multicolumn{2}{c}{\textbf{Annotators}} &
   \multicolumn{2}{c}{\textbf{Ours}} \\
\cmidrule(lr){2-3} \cmidrule(lr){4-5}
\textbf{Event Type} &
  Fwd (s) & Bwd (s) & Fwd (s) & Bwd (s) \\
\midrule
Save         & 1.8$\pm$0.6  & 3.8$\pm$0.3  & 2  & 4  \\
Dribble      & 3.9$\pm$1.3  & 2.9$\pm$0.8  & 4  & 3  \\
Tackle       & 4.2$\pm$2.0  & 2.6$\pm$0.7  & 5  & 3  \\
Block        & 4.2$\pm$2.2  & 3.4$\pm$1.0  & 5  & 3  \\
Clearance    & 4.3$\pm$1.6  & 3.3$\pm$1.0  & 5  & 4  \\
Shot         & 4.3$\pm$1.8  & 4.0$\pm$2.7  & 4  & 8  \\
Foul         & 4.4$\pm$1.0  & 7.3$\pm$2.7  & 5  & 7  \\
Yellow Card  & 8.9$\pm$2.6  & 22.8$\pm$0.4 & 10 & 23 \\
Substitution & 15.9$\pm$3.5 & 12.8$\pm$2.7 & 10 & 12 \\
\bottomrule
\end{tabular}%
}
\end{table}

\section{MLLM Experiment Details}
\label{sec:mllm_details}

\subsection{Inference Prompts}
\label{sec:mllm_prompts}

We evaluate two categories of MLLMs on the GMR task: general-purpose
models (Qwen3-VL-4B/8B/32B)~\cite{bai2025qwen3} and temporal grounding specialists
(TRACE~\cite{guo2024trace}, TimeChat~\cite{ren2024timechat}). Below we provide the exact inference prompts used
for each category. The GRPO-fine-tuned Qwen3-VL-4B uses the same
prompt as the zero-shot Qwen3-VL variants.

\begin{tcolorbox}[colback=gray!5, colframe=gray!60,
  fontupper=\small\ttfamily,
  boxsep=2pt, top=2pt, bottom=2pt, left=4pt, right=4pt,
  title={\small\sffamily Prompt for General-Purpose MLLMs (Qwen3-VL)}]
Locate all relevant time windows in the video
based on the text query.\\
Respond with JSON only, no extra text.\\
Output format:\\
\{"relevant\_windows": [[start, end], ...]\}\\
Note: start/end must be numeric seconds (decimals
allowed). Do NOT use formats like 0:38 or
00:38:12.\\
If the queried event does not exist, output:\\
\{"relevant\_windows": []\}\\[4pt]
Video duration: \{duration\} seconds.\\
Query: \{query\}
\end{tcolorbox}

\begin{tcolorbox}[colback=gray!5, colframe=gray!60,
  fontupper=\small\ttfamily,
  boxsep=2pt, top=2pt, bottom=2pt, left=4pt, right=4pt,
  title={\small\sffamily Prompt for Grounding MLLMs (TRACE, TimeChat)}]
Localize all visual content described by the given
textual query '\{query\}' in the video, and output
the start and end timestamps in seconds for each
occurrence. The event may occur multiple times or
may not exist at all. If the event does not exist
in the video, do not output any timestamps.
\end{tcolorbox}

\noindent
Both prompts explicitly instruct the model to handle multi-moment
retrieval and null-set rejection, the two core challenges of GMR.
The general-purpose prompt enforces structured JSON output for
reliable parsing, while the grounding prompt follows each model's
native interface with added GMR-specific instructions.

\subsection{Reward Function Design}
\label{sec:grpo_reward}

We design a composite reward function for Group Relative Policy Optimization (GRPO)~\cite{shao2024deepseekmath}
that provides dense, structured supervision to guide the multimodal LLM toward accurate
moment retrieval outputs. The reward consists of two components: a \emph{format reward}
$r_{\text{fmt}}$ and a \emph{content reward} $r_{\text{cont}}$.
We denote the KL penalty weight by $\beta_{\mathrm{KL}}$ (Tab.~\ref{tab:grpo_hyper})
to avoid confusion with the negative-to-positive ratio $\beta$ in
Alg.~\ref{alg:balanced_sampling}.

\paragraph{Format Reward.}
The format reward provides graduated feedback on output structure compliance.
Given model output $\hat{y}$, we define:
\begin{equation}
  r_{\text{fmt}}(\hat{y}) =
  \begin{cases}
    \phantom{-}0.0  & \text{valid \texttt{<answer>} tags with well-formed JSON}, \\
    -0.2 & \text{valid tags, malformed JSON payload}, \\
    -0.3 & \text{regex match but corrupted content}, \\
    -0.5 & \text{opening \texttt{<answer>} tag only (truncated)}, \\
    -1.0 & \text{no recognizable tags}.
  \end{cases}
  \label{eq:format_reward}
\end{equation}

\paragraph{Content Reward.}
Let $\mathcal{G} = \{g_1, \dots, g_M\}$ denote the set of ground-truth windows and
$\mathcal{P} = \{p_1, \dots, p_N\}$ the predicted windows extracted from $\hat{y}$.
We define the content reward as follows.

\smallskip\noindent\textbf{Case 1: Null ground truth} ($M = 0$).
\begin{equation}
  r_{\text{cont}} =
  \begin{cases}
    +0.1 & \text{if } N = 0 \;\text{(correct rejection)}, \\
    -0.3 - 0.1 \cdot \min(N, N_{\max}) & \text{if } N > 0 \;\text{(false positive)}.
  \end{cases}
  \label{eq:null_gt}
\end{equation}

\smallskip\noindent\textbf{Case 2: Non-empty ground truth, empty prediction}
($M > 0, N = 0$).
\begin{equation}
  r_{\text{cont}} = -0.7.
  \label{eq:empty_pred}
\end{equation}

\smallskip\noindent\textbf{Case 3: Non-empty ground truth and prediction}
($M > 0, N > 0$).
For each $k \in \{1, 2, 3\}$, let $k' = \min(k, N)$. We compute
per-sample recall $\mRatk$~\cite{lei2021detecting} averaged over IoU
thresholds $\Theta=\{0.50,0.55,\ldots,0.95\}$
via greedy bipartite matching:
\begin{equation}
  \mRatk = \frac{1}{|\Theta|} \sum_{\theta \in \Theta}
    \frac{\bigl|\text{Match}(\mathcal{P}_{:k'}, \mathcal{G}, \theta)\bigr|}{M},
  \label{eq:mr_at_k}
\end{equation}
where $\text{Match}(\cdot, \cdot, \theta)$ performs greedy bipartite matching
with IoU threshold $\theta$.
Similarly, we compute per-sample $\mIoUatk$ by forcing all matches
(threshold $\theta = -1$) and averaging matched IoU values:
\begin{equation}
  \mIoUatk = \frac{1}{M}
    \sum_{(p_i, g_j) \in \text{Match}(\mathcal{P}_{:k'}, \mathcal{G}, -1)}
    \text{tIoU}(p_i, g_j).
  \label{eq:miou_at_k}
\end{equation}

An overlap bonus encourages coarse localization even when IoU is low:
\begin{equation}
  r_{\text{overlap}} = 0.15 \cdot
    \frac{\bigl|\text{Match}(\mathcal{P}, \mathcal{G}, 0.01)\bigr|}{M}.
  \label{eq:overlap}
\end{equation}

Let $n_{\mathrm{zt}}$ be the number of zero-length windows after clipping,
$n_{\mathrm{dur}}$ the number of predicted windows whose endpoints exceed
the video duration (after clipping), and
$n_{\mathrm{exc}} = \max(0,\, N - N_{\max})$ the number of windows beyond
$N_{\max} = 10$. We define the validity penalty as
\begin{equation}
  r_{\text{penalty}} =
    -0.2\, n_{\mathrm{zt}} - 0.05\, n_{\mathrm{dur}} - 0.1\, n_{\mathrm{exc}}.
  \label{eq:penalty}
\end{equation}

The content reward for Case~3 is:
\begin{equation}
  r_{\text{cont}} = \sum_{k=1}^{3}
    \bigl(w_k^{\mathrm{mR}}\,\mRatk + w_k^{\mathrm{mIoU}}\,\mIoUatk\bigr)
    + r_{\text{overlap}} + r_{\text{penalty}},
  \label{eq:content_reward}
\end{equation}
with weights $\mathbf{w}^{\mathrm{mR}} = (0.45, 0.35, 0.20)$ and
$\mathbf{w}^{\mathrm{mIoU}} = (0.20, 0.15, 0.10)$.

\paragraph{Final Reward.}
The total reward is:
\begin{equation}
  r = r_{\text{cont}} + w_{\text{fmt}} \cdot r_{\text{fmt}},
  \label{eq:total_reward}
\end{equation}
with $w_{\text{fmt}} = 0.3$, clipped to $[-1, 1]$.
When parsing fails entirely ($\mathcal{P} = \texttt{null}$), we fall back to
$r = w_{\text{fmt}} \cdot r_{\text{fmt}} + (1 - w_{\text{fmt}}) \cdot r_{\text{fail}}$,
where $r_{\text{fail}} = -1.0$ ensures that even unparseable outputs receive
a gradient signal from the format component.

\subsection{Training Configuration}
\label{sec:grpo_training}

We fine-tune Qwen3-VL-4B-Instruct using LoRA~\cite{hu2022lora} with
rank $r = 16$, scaling factor $\alpha = 32$, and dropout $0.05$,
applied to all linear layers of the language model while keeping
the vision encoder and aligner frozen.
Training uses the GRPO objective~\cite{shao2024deepseekmath} with the reward function described in
Sec.~\ref{sec:grpo_reward}.

\begin{table}[h]
\centering
\caption{GRPO training hyperparameters.}
\label{tab:grpo_hyper}
\small
\begin{tabular}{ll}
\toprule
\textbf{Hyperparameter} & \textbf{Value} \\
\midrule
Base model            & Qwen3-VL-4B-Instruct \\
LoRA rank / $\alpha$  & 16 / 32 \\
LoRA target modules   & all-linear \\
LoRA dropout          & 0.05 \\
Optimizer             & AdamW (fused) \\
$(\beta_1, \beta_2)$  & (0.9, 0.95) \\
Learning rate         & $1 \times 10^{-5}$ (cosine) \\
Weight decay          & 0.1 \\
Max gradient norm     & 1.0 \\
Precision             & BF16 \\
Batch size / GPU      & 1 \\
Gradient accumulation & 4 \\
Number of GPUs        & 3 \\
Effective batch size  & 12 \\
Epochs                & 1 \\
Num.\ generations ($G$) & 4 \\
GRPO $\epsilon$ (clip) & 0.2 \\
KL coefficient $\beta_{\mathrm{KL}}$ & 0.04 \\
Temperature           & 0.9 \\
Top-$p$ / Top-$k$     & 0.9 / 50 \\
Max sequence length   & 8192 \\
Max completion length & 1024 \\
Seed                  & 42 \\
Gradient checkpointing & \checkmark \\
\bottomrule
\end{tabular}
\end{table}

\section{Query Style Robustness Details}
\label{sec:query_style}

This section supplements the query style robustness analysis in
Sec.~5.4 of the main paper. All five reformulations preserve the
same core semantic content (event type and attribute constraints),
differing only in surface form and length.
We group them into \emph{phrasing variants} (B, C), which alter
sentence structure at comparable lengths, and \emph{length variants}
(D, E), which substantially shorten or lengthen the query.
Below, we describe the construction rule for each style.

\paragraph{Original (Baseline, $\sim$7 words).}
The original queries are the base queries present in the
Soccer-GMR dataset. Each query expresses the target
event-attribute semantics in a concise imperative sentence,
e.g., \textit{"Locate all shot actions by players from Canada."}

\paragraph{B: Question (Phrasing Variant, $\sim$9 words).}
The original imperative sentence is converted into an interrogative
form. The verb phrase is restructured using a \textit{wh-question} 
word (typically \textit{"}when"), and the attribute
clause is repositioned as the subject or modifier.
E.g., \textit{"When did Canadian players perform a shot?"}

\paragraph{C: Noun Phrase (Phrasing Variant, $\sim$8 words).}
The imperative verb (\textit{Locate}, \textit{Find}, etc.) is
removed, and the remaining content is reformulated as a nominal
expression with the event type as the head noun and attributes
expressed as post-modifiers.
E.g., \textit{"A shot performed by Canadian players."}

\paragraph{D: Keyword (Length Variant, $\sim$3 words).}
All function words, verbs, and syntactic structures are discarded.
Only the event type and key attribute words are retained in their
bare form, producing a minimal keyword-style query.
E.g., \textit{"Canada shot"}

\paragraph{E: Verbose (Length Variant, $\sim$28 words).}
Detailed task instructions are prepended to the original query,
explicitly directing the model to examine the entire video and
retrieve all matching moments. The core semantic content remains
unchanged; only the surrounding instructional context is added.
E.g., \textit{"Please go through the entire video carefully and
locate all shot actions performed by players from Canada."}

\smallskip
Table~\ref{tab:query_style_examples} provides a side-by-side
comparison. Results are reported in Table~3 of the
main paper.

\begin{table}[h]
\centering
\caption{Query style reformulation examples. All variants are
derived from the same base event
\texttt{(shot, by players from Canada)}.}
\label{tab:query_style_examples}
\small
\resizebox{\columnwidth}{!}{%
\begin{tabular}{lcp{6.8cm}}
\toprule
\textbf{Style} & \textbf{Avg.\ Len.} & \textbf{Example} \\
\midrule
Original   & $\sim$7  & Locate all shot actions by players from Canada. \\
B: Question    & $\sim$9  & When did Canada players perform a shot? \\
C: Noun Phrase & $\sim$8  & A shot performed by Canada players. \\
D: Keyword     & $\sim$3  & Canada shot \\
E: Verbose     & $\sim$28 & Please go through the entire video carefully and locate all shot actions performed by players from Canada. \\
\bottomrule
\end{tabular}%
}
\end{table}

\section{Metric Threshold Sensitivity}
\label{sec:sensitivity}

Table~\ref{tab:threshold_sweep} reports G-mIoU@1 and Rej-F1 at three
operating thresholds $\tau \in \{0.4, 0.6, 0.8\}$, along with the
average across thresholds (AP).
For base models without an explicit existence score, we use
$\max(\text{window score})$ as a proxy; GMR variants use the dedicated
\texttt{pred\_exist\_score}.

\begin{table}[h]
\centering
\caption{Threshold sensitivity of G-mIoU@1 and Rej-F1. AP denotes
the average across thresholds.
\textbf{Bold}: best; \underline{underline}: second best.}
\label{tab:threshold_sweep}
\small
\resizebox{\columnwidth}{!}{%
\begin{tabular}{l cccc cccc}
\toprule
 & \multicolumn{4}{c}{\textbf{G-mIoU@1}} &
   \multicolumn{4}{c}{\textbf{Rej-F1}} \\
\cmidrule(lr){2-5} \cmidrule(lr){6-9}
\textbf{Model} &
  $\tau$=0.4 & $\tau$=0.6 & $\tau$=0.8 & AP &
  $\tau$=0.4 & $\tau$=0.6 & $\tau$=0.8 & AP \\
\midrule
Moment-DETR   & 5.39  & 5.39  & 6.04  & 5.52  & 0.00  & 0.00  & 2.78  & 0.56 \\
QD-DETR       & 8.48  & 8.57  & 10.20 & 8.88  & 0.00  & 0.40  & 7.25  & 1.77 \\
CG-DETR       & 11.83 & 11.83 & 11.93 & 11.85 & 0.00  & 0.00  & 0.40  & 0.08 \\
EaTR          & 12.94 & 13.64 & 17.35 & 14.28 & 0.80  & 3.95  & 17.76 & 6.21 \\
FlashVTG      & 15.41 & 47.49 & 47.49 & 38.96 & 7.12  & 64.40 & 64.40 & 51.16 \\
\midrule
Moment-DETR-GMR & 35.84 & 51.64 & 51.64 & \underline{48.26} & 64.01 & 75.00 & 75.00 & \textbf{72.70} \\
EaTR-GMR      & 37.89 & 45.84 & 51.45 & 45.01 & 62.10 & 71.05 & 74.22 & 69.33 \\
FlashVTG-GMR  & 39.58 & 51.38 & 54.13 & \textbf{49.43} & 61.72 & 73.06 & 74.63 & \underline{70.94} \\
\bottomrule
\end{tabular}%
}
\end{table}

\paragraph{Observations.}
(1)~The relative ranking among models remains consistent across all
tested thresholds, demonstrating that the benchmark conclusions in
the main paper are robust to the choice of~$\tau$.
(2)~GMR variants consistently outperform their base counterparts by
a large margin on both metrics, confirming that base models possess
limited rejection capability under the GMR setting.
(3)~FlashVTG-GMR achieves the highest AP(G-mIoU@1) despite slightly
lower AP(Rej-F1) compared to Moment-DETR-GMR, indicating that its
stronger localization quality compensates for relatively weaker
rejection; this highlights the value of G-mIoU as a joint metric
that captures both abilities simultaneously.

\begin{table*}[htbp]
\centering
\caption{Data Statistics of Gymnastics-GMR. $^\star$\,Duration-flexible instantiation: \SI{300}{\second} windows here versus
\SI{150}{\second} in Soccer-GMR.}
\label{tab:gymnastics_gmr_comparison}
\small
\begin{tabular}{l c c c c c c c c}
\toprule
\textbf{Dataset} & \textbf{Domain} & \textbf{\# Queries} &
\textbf{\begin{tabular}[c]{@{}c@{}}\# Moments /\\ \# Videos\end{tabular}} &
\textbf{\begin{tabular}[c]{@{}c@{}}Avg.\ Moment /\\ Query w/ Target\end{tabular}} &
\textbf{\begin{tabular}[c]{@{}c@{}}Avg.\ Video\\ Dur.\end{tabular}} &
\textbf{\begin{tabular}[c]{@{}c@{}}Multi-\\ Moment\end{tabular}} &
\textbf{\begin{tabular}[c]{@{}c@{}}Null-\\ Set\end{tabular}} &
\textbf{\begin{tabular}[c]{@{}c@{}}Duration\\ Flexible\end{tabular}} \\
\midrule
Gymnastics-GMR & Gymnastics &
\num{2012} & \num{1.5}K / \num{176} &
\num{1.5} & \SI{300}{\second}$^\star$ &
\checkmark & \checkmark & \checkmark \\
\bottomrule
\end{tabular}
\end{table*}

\section{Additional Domain Instantiation}
\label{sec:cross_domain_finegym}

\textbf{Gymnastics-GMR} applies the Sec.~3.2 construction pipeline to FineGym's Gym99
hierarchy~\cite{shao2020finegym}, providing another domain in which
the same stages yield a structured GMR split.
Stage~II uses \SI{300}{\second} windows with \SI{30}{\second} overlap, whereas our
Soccer-GMR build uses \SI{150}{\second} with \SI{10}{\second} overlap, illustrating
that clip duration remains a configurable instantiation of the duration-flexible design
in Sec.~3.2.

We take the first \num{2000} Gym99 \emph{Val} element list lines, merge contiguous
mentions, and obtain \num{509} query identities over \num{8} source videos.
\textbf{Stage~I} forms structured natural-language queries from FineGym's captions and segments metadata, parallel to
$\langle\text{event},\text{attributes}\rangle$ queries in Sec.~3.2.
\textbf{Stage~II} performs sliding-window clipping and balanced sampling as in Sec.~3.2.
\textbf{Stage~III} applies the same query diversification.

\begin{figure}[t]
  \centering
  \includegraphics[width=\linewidth]{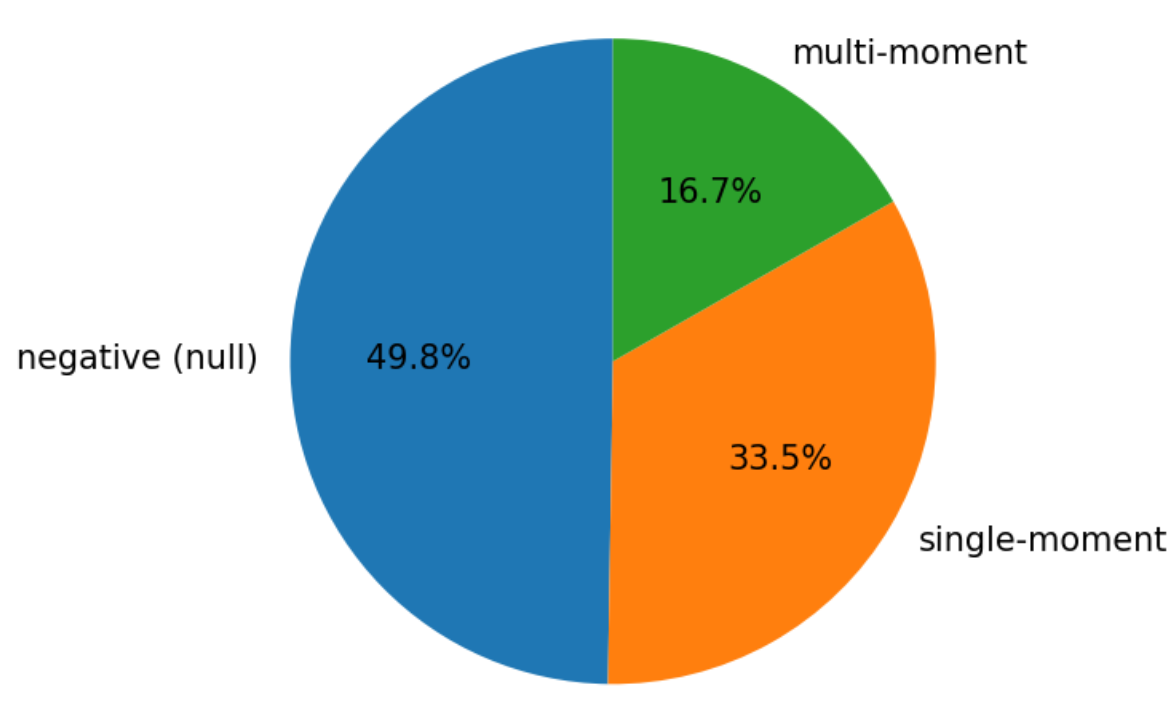}
  \caption{Gymnastics-GMR: Query types mix after balancing (null- vs.\ single- vs.\ multi-moment).}
  \label{fig:gymnastics_row_mix}
\end{figure}

\begin{figure}[t]
  \centering
  \includegraphics[width=\linewidth]{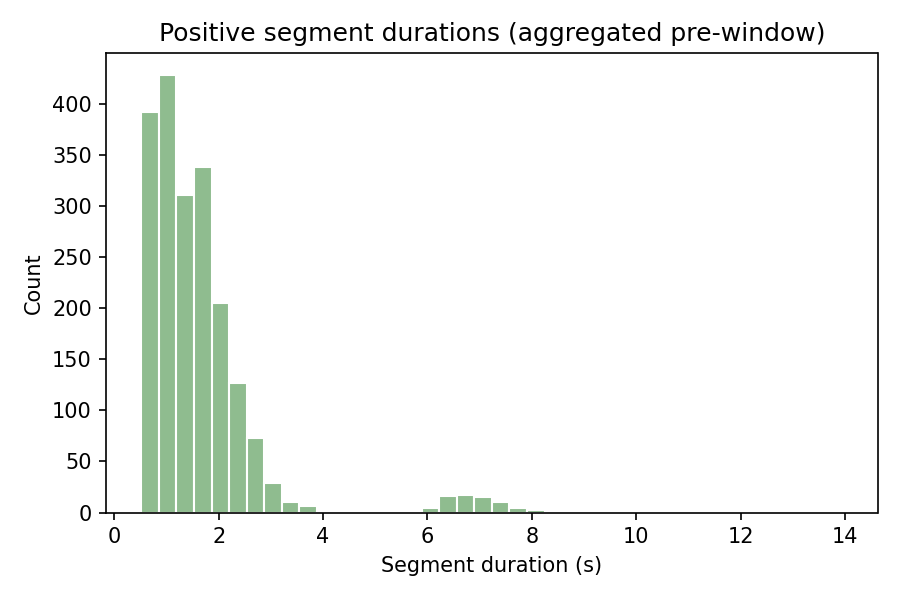}
  \caption{Gymnastics-GMR: positive segment durations in aggregated annotations prior to sliding windows.}
  \label{fig:gymnastics_seg_hist}
\end{figure}

Table~\ref{tab:gymnastics_gmr_comparison} summarizes Gymnastics-GMR alongside the
metrics used in the main paper's dataset comparison.
The split includes null-set, single-moment, and multi-moment rows with a near
\num{2}:\num{1} single-to-multi ratio among positives and \num{1502} ground-truth
segments on positive queries.
Figures~\ref{fig:gymnastics_row_mix}--\ref{fig:gymnastics_seg_hist} visualize the label types, and segment-length behavior.

\section{Benchmark Split and Released Data}
\label{sec:benchmark_split}

In our experiments, we use a fixed Soccer-GMR benchmark split containing 1,957 video clips and 5,639 query-moment pairs, including 2,935 positive samples and 2,704 null-set samples. Among the positive samples, 1,972 contain a single ground-truth moment and 963 contain multiple moments.

The split is constructed at the video-clip level, with no video clip shared across train, validation, and test. Query-moment pairs inherit the split of their source clips, avoiding duplicate visual content across training and evaluation. In addition to this benchmark split, we publicly release the full Soccer-GMR dataset, including all 22,119 query-moment pairs, to support future research on generalized moment retrieval and larger-scale model training.

\section{Dataset Release Plan}
\label{sec:release_plan}
All data and code are publicly available at
\url{https://github.com/dymm9977/generalized-moment-retrieval}
to facilitate future research on generalized moment retrieval.

\bibliographystyle{ACM-Reference-Format}
\bibliography{sample-base}

@String{Computing = "Computing" }

@String{Computer = "{IEEE} Computer" }

@String{Springer = "Springer-Verlag" }

@ArtifactSoftware{R,
    title = {R: A Language and Environment for Statistical Computing},
    author = {{R Core Team}},
    organization = {R Foundation for Statistical Computing},
    address = {Vienna, Austria},
    year = {2019},
    url = {https://www.R-project.org/},
}

@inproceedings{chen2025grounded,
  title={Grounded multi-hop videoqa in long-form egocentric videos},
  author={Chen, Qirui and Di, Shangzhe and Xie, Weidi},
  booktitle={Proceedings of the AAAI Conference on Artificial Intelligence},
  volume={39},
  number={2},
  pages={2159--2167},
  year={2025}
}

@article{bai2025qwen3,
  title={Qwen3-vl technical report},
  author={Bai, Shuai and Cai, Yuxuan and Chen, Ruizhe and Chen, Keqin and Chen, Xionghui and Cheng, Zesen and Deng, Lianghao and Ding, Wei and Gao, Chang and Ge, Chunjiang and others},
  journal={arXiv preprint arXiv:2511.21631},
  year={2025}
}

@article{zhang2025videollama,
  title={Videollama 3: Frontier multimodal foundation models for image and video understanding},
  author={Zhang, Boqiang and Li, Kehan and Cheng, Zesen and Hu, Zhiqiang and Yuan, Yuqian and Chen, Guanzheng and Leng, Sicong and Jiang, Yuming and Zhang, Hang and Li, Xin and others},
  journal={arXiv preprint arXiv:2501.13106},
  year={2025}
}

@article{zhang2024llava,
  title={Llava-video: Video instruction tuning with synthetic data},
  author={Zhang, Yuanhan and Wu, Jinming and Li, Wei and Li, Bo and Ma, Zejun and Liu, Ziwei and Li, Chunyuan},
  journal={arXiv preprint arXiv:2410.02713},
  year={2024}
}

@inproceedings{abdessaied2025v,
  title={V\^{} 2Dial: Unification of Video and Visual Dialog via Multimodal Experts},
  author={Abdessaied, Adnen and Rohrbach, Anna and Rohrbach, Marcus and Bulling, Andreas},
  booktitle={Proceedings of the Computer Vision and Pattern Recognition Conference},
  pages={8637--8647},
  year={2025}
}

@inproceedings{abdessaied2024multi,
  title={Multi-modal video dialog state tracking in the wild},
  author={Abdessaied, Adnen and Shi, Lei and Bulling, Andreas},
  booktitle={European Conference on Computer Vision},
  pages={348--365},
  year={2024},
  organization={Springer}
}

@inproceedings{zhang2025bridging,
  title={Bridging modalities: Improving universal multimodal retrieval by multimodal large language models},
  author={Zhang, Xin and Zhang, Yanzhao and Xie, Wen and Li, Mingxin and Dai, Ziqi and Long, Dingkun and Xie, Pengjun and Zhang, Meishan and Li, Wenjie and Zhang, Min},
  booktitle={Proceedings of the Computer Vision and Pattern Recognition Conference},
  pages={9274--9285},
  year={2025}
}

@article{lee2025generalized,
  title={Generalized contrastive learning for universal multimodal retrieval},
  author={Lee, Jungsoo and Cho, Janghoon and Park, Hyojin and Hayat, Munawar and Hwang, Kyuwoong and Porikli, Fatih and Choi, Sungha},
  journal={arXiv preprint arXiv:2509.25638},
  year={2025}
}

@inproceedings{xing2025context,
  title={Context-cir: Learning from concepts in text for composed image retrieval},
  author={Xing, Eric and Kolouju, Pranavi and Pless, Robert and Stylianou, Abby and Jacobs, Nathan},
  booktitle={Proceedings of the Computer Vision and Pattern Recognition Conference},
  pages={19638--19648},
  year={2025}
}

@inproceedings{deng2025motion,
  title={Motion-grounded video reasoning: Understanding and perceiving motion at pixel level},
  author={Deng, Andong and Chen, Tongjia and Yu, Shoubin and Yang, Taojiannan and Spencer, Lincoln and Tian, Yapeng and Mian, Ajmal Saeed and Bansal, Mohit and Chen, Chen},
  booktitle={Proceedings of the Computer Vision and Pattern Recognition Conference},
  pages={8625--8636},
  year={2025}
}

@inproceedings{liu2025commonsense,
  title={Commonsense video question answering through video-grounded entailment tree reasoning},
  author={Liu, Huabin and Ilievski, Filip and Snoek, Cees GM},
  booktitle={Proceedings of the Computer Vision and Pattern Recognition Conference},
  pages={3262--3271},
  year={2025}
}

@inproceedings{chen2025cross,
  title={Cross-modal causal relation alignment for video question grounding},
  author={Chen, Weixing and Liu, Yang and Chen, Binglin and Su, Jiandong and Zheng, Yongsen and Lin, Liang},
  booktitle={Proceedings of the Computer Vision and Pattern Recognition Conference},
  pages={24087--24096},
  year={2025}
}

@article{chen2024verified,
  title={Verified: A video corpus moment retrieval benchmark for fine-grained video understanding},
  author={Chen, Houlun and Wang, Xin and Chen, Hong and Zhang, Zeyang and Feng, Wei and Huang, Bin and Jia, Jia and Zhu, Wenwu},
  journal={Advances in Neural Information Processing Systems},
  volume={37},
  pages={40393--40406},
  year={2024}
}

@inproceedings{liang2025tvr,
  title={Tvr-ranking: A dataset for ranked video moment retrieval with imprecise queries},
  author={Liang, Renjie and Zhang, Chongzhi and Li, Li and Wang, Jing and Zhu, Xizhou and Sun, Aixin},
  booktitle={Proceedings of the 2025 Annual International ACM SIGIR Conference on Research and Development in Information Retrieval in the Asia Pacific Region},
  pages={231--239},
  year={2025}
}

@article{zhao2025ld,
  title={Ld-detr: Loop decoder detection transformer for video moment retrieval and highlight detection},
  author={Zhao, Pengcheng and He, Zhixian and Zhang, Fuwei and Lin, Shujin and Zhou, Fan},
  journal={arXiv preprint arXiv:2501.10787},
  year={2025}
}

@article{lei2021detecting,
  title={Detecting moments and highlights in videos via natural language queries},
  author={Lei, Jie and Berg, Tamara L and Bansal, Mohit},
  journal={Advances in Neural Information Processing Systems},
  volume={34},
  pages={11846--11858},
  year={2021}
}

@inproceedings{ma2025ms,
  title={Ms-detr: Towards effective video moment retrieval and highlight detection by joint motion-semantic learning},
  author={Ma, Hongxu and Wang, Guanshuo and Yu, Fufu and Jia, Qiong and Ding, Shouhong},
  booktitle={Proceedings of the 33rd ACM International Conference on Multimedia},
  pages={4514--4523},
  year={2025}
}

@article{cao2025one,
  title={When One Moment Isn't Enough: Multi-Moment Retrieval with Cross-Moment Interactions},
  author={Cao, Zhuo and Du, Heming and Zhang, Bingqing and Yu, Xin and Li, Xue and Wang, Sen},
  journal={arXiv preprint arXiv:2510.17218},
  year={2025}
}

@inproceedings{yang2024new,
  title={A New Framework for Evaluating Faithfulness of Video Moment Retrieval against Multiple Distractors},
  author={Yang, Nakyeong and Kim, Minsung and Yoon, Seunghyun and Shin, Joongbo and Jung, Kyomin},
  booktitle={Proceedings of the 33rd ACM International Conference on Information and Knowledge Management},
  pages={2869--2878},
  year={2024}
}

@article{chen2025prvr,
  title={Prvr: Partially relevant video retrieval},
  author={Chen, Xianke and Liu, Daizong and Yang, Xun and Li, Xirong and Dong, Jianfeng and Wang, Meng and Wang, Xun},
  journal={IEEE Transactions on Pattern Analysis and Machine Intelligence},
  year={2025},
  publisher={IEEE}
}

@inproceedings{flanagan2025moment,
  title={Moment of Untruth: Dealing with Negative Queries in Video Moment Retrieval},
  author={Flanagan, Kevin and Damen, Dima and Wray, Michael},
  booktitle={2025 IEEE/CVF Winter Conference on Applications of Computer Vision (WACV)},
  pages={5336--5345},
  year={2025},
  organization={IEEE}
}

@inproceedings{moon2023query,
  title={Query-dependent video representation for moment retrieval and highlight detection},
  author={Moon, WonJun and Hyun, Sangeek and Park, SangUk and Park, Dongchan and Heo, Jae-Pil},
  booktitle={Proceedings of the IEEE/CVF conference on computer vision and pattern recognition},
  pages={23023--23033},
  year={2023}
}

@inproceedings{li2022compositional,
  title={Compositional temporal grounding with structured variational cross-graph correspondence learning},
  author={Li, Juncheng and Xie, Junlin and Qian, Long and Zhu, Linchao and Tang, Siliang and Wu, Fei and Yang, Yi and Zhuang, Yueting and Wang, Xin Eric},
  booktitle={Proceedings of the IEEE/CVF Conference on Computer Vision and Pattern Recognition},
  pages={3032--3041},
  year={2022}
}

@article{shao2024deepseekmath,
  title={Deepseekmath: Pushing the limits of mathematical reasoning in open language models},
  author={Shao, Zhihong and Wang, Peiyi and Zhu, Qihao and Xu, Runxin and Song, Junxiao and Bi, Xiao and Zhang, Haowei and Zhang, Mingchuan and Li, YK and Wu, Yang and others},
  journal={arXiv preprint arXiv:2402.03300},
  year={2024}
}

@article{liu2023survey,
  title={A survey on video moment localization},
  author={Liu, Meng and Nie, Liqiang and Wang, Yunxiao and Wang, Meng and Rui, Yong},
  journal={ACM Computing Surveys},
  volume={55},
  number={9},
  pages={1--37},
  year={2023},
  publisher={ACM New York, NY}
}

@article{lan2023survey,
  title={A survey on temporal sentence grounding in videos},
  author={Lan, Xiaohan and Yuan, Yitian and Wang, Xin and Wang, Zhi and Zhu, Wenwu},
  journal={ACM Transactions on Multimedia Computing, Communications and Applications},
  volume={19},
  number={2},
  pages={1--33},
  year={2023},
  publisher={ACM New York, NY}
}

@inproceedings{woo2024let,
  title={Let me finish my sentence: Video temporal grounding with holistic text understanding},
  author={Woo, Jongbhin and Ryu, Hyeonggon and Jang, Youngjoon and Cho, Jae Won and Chung, Joon Son},
  booktitle={Proceedings of the 32nd ACM International Conference on Multimedia},
  pages={8199--8208},
  year={2024}
}

@inproceedings{jang2023knowing,
  title={Knowing where to focus: Event-aware transformer for video grounding},
  author={Jang, Jinhyun and Park, Jungin and Kim, Jin and Kwon, Hyeongjun and Sohn, Kwanghoon},
  booktitle={Proceedings of the IEEE/CVF International Conference on Computer Vision},
  pages={13846--13856},
  year={2023}
}

@article{moon2023correlation,
  title={Correlation-guided query-dependency calibration for video temporal grounding},
  author={Moon, WonJun and Hyun, Sangeek and Lee, SuBeen and Heo, Jae-Pil},
  journal={arXiv preprint arXiv:2311.08835},
  year={2023}
}

@inproceedings{cao2025flashvtg,
  title={Flashvtg: Feature layering and adaptive score handling network for video temporal grounding},
  author={Cao, Zhuo and Zhang, Bingqing and Du, Heming and Yu, Xin and Li, Xue and Wang, Sen},
  booktitle={2025 IEEE/CVF Winter Conference on Applications of Computer Vision (WACV)},
  pages={9226--9236},
  year={2025},
  organization={IEEE}
}

@article{wu2025survey,
  title={A survey on video temporal grounding with multimodal large language model},
  author={Wu, Jianlong and Liu, Wei and Liu, Ye and Liu, Meng and Nie, Liqiang and Lin, Zhouchen and Chen, Chang Wen},
  journal={IEEE Transactions on Pattern Analysis and Machine Intelligence},
  year={2025},
  publisher={IEEE}
}

@inproceedings{wang2026spacevllm,
  title={Spacevllm: Endowing multimodal large language model with spatio-temporal video grounding capability},
  author={Wang, Jiankang and Zhang, Zhihan and Liu, Zhihang and Li, Yang and Ge, Jiannan and Xie, Hongtao and Zhang, Yongdong},
  booktitle={Proceedings of the AAAI Conference on Artificial Intelligence},
  volume={40},
  number={12},
  pages={9912--9920},
  year={2026}
}

@inproceedings{pramanick2025enrich,
  title={Enrich and Detect: Video Temporal Grounding with Multimodal LLMs},
  author={Pramanick, Shraman and Mavroudi, Effrosyni and Song, Yale and Chellappa, Rama and Torresani, Lorenzo and Afouras, Triantafyllos},
  booktitle={Proceedings of the IEEE/CVF International Conference on Computer Vision},
  pages={24297--24308},
  year={2025}
}

@inproceedings{fang2024not,
  title={Not all inputs are valid: Towards open-set video moment retrieval using language},
  author={Fang, Xiang and Fang, Wanlong and Liu, Daizong and Qu, Xiaoye and Dong, Jianfeng and Zhou, Pan and Li, Renfu and Xu, Zichuan and Chen, Lixing and Zheng, Panpan and others},
  booktitle={Proceedings of the 32nd ACM International Conference on Multimedia},
  pages={28--37},
  year={2024}
}

@inproceedings{krishna2017dense,
  title={Dense-captioning events in videos},
  author={Krishna, Ranjay and Hata, Kenji and Ren, Frederic and Fei-Fei, Li and Carlos Niebles, Juan},
  booktitle={Proceedings of the IEEE international conference on computer vision},
  pages={706--715},
  year={2017}
}

@inproceedings{deliege2021soccernet,
  title={Soccernet-v2: A dataset and benchmarks for holistic understanding of broadcast soccer videos},
  author={Deliege, Adrien and Cioppa, Anthony and Giancola, Silvio and Seikavandi, Meisam J and Dueholm, Jacob V and Nasrollahi, Kamal and Ghanem, Bernard and Moeslund, Thomas B and Van Droogenbroeck, Marc},
  booktitle={Proceedings of the IEEE/CVF conference on computer vision and pattern recognition},
  pages={4508--4519},
  year={2021}
}

@inproceedings{qin2025generalized,
  title={Generalized video moment retrieval},
  author={Qin, You and Wu, Qilong and Li, Yicong and Ji, Wei and Li, Li and Cai, Pengcheng and Wei, Lina and Zimmermann, Roger},
  booktitle={The Thirteenth International Conference on Learning Representations},
  year={2025}
}

@inproceedings{gao2017tall,
  title={TALL: Temporal Activity Localization via Language Query},
  author={Gao, Jiyang and Sun, Chen and Yang, Zhenheng and Nevatia, Ram},
  booktitle={ICCV},
  year={2017}
}

@inproceedings{lei2020tvr,
  title={Tvr: A large-scale dataset for video-subtitle moment retrieval},
  author={Lei, Jie and Yu, Licheng and Berg, Tamara L and Bansal, Mohit},
  booktitle={European Conference on Computer Vision},
  pages={447--463},
  year={2020},
  organization={Springer}
}

@inproceedings{anne2017localizing,
  title={Localizing moments in video with natural language},
  author={Anne Hendricks, Lisa and Wang, Oliver and Shechtman, Eli and Sivic, Josef and Darrell, Trevor and Russell, Bryan},
  booktitle={Proceedings of the IEEE international conference on computer vision},
  pages={5803--5812},
  year={2017}
}

@article{regneri2013grounding,
  title={Grounding action descriptions in videos},
  author={Regneri, Michaela and Rohrbach, Marcus and Wetzel, Dominikus and Thater, Stefan and Schiele, Bernt and Pinkal, Manfred},
  journal={Transactions of the Association for Computational Linguistics},
  volume={1},
  pages={25--36},
  year={2013},
  publisher={MIT Press One Rogers Street, Cambridge, MA 02142-1209, USA journals-info~…}
}

@inproceedings{ren2024timechat,
  title={Timechat: A time-sensitive multimodal large language model for long video understanding},
  author={Ren, Shuhuai and Yao, Linli and Li, Shicheng and Sun, Xu and Hou, Lu},
  booktitle={Proceedings of the IEEE/CVF Conference on Computer Vision and Pattern Recognition},
  pages={14313--14323},
  year={2024}
}

@article{guo2024trace,
  title={Trace: Temporal grounding video llm via causal event modeling},
  author={Guo, Yongxin and Liu, Jingyu and Li, Mingda and Liu, Qingbin and Chen, Xi and Tang, Xiaoying},
  journal={arXiv preprint arXiv:2410.05643},
  year={2024}
}

@article{zhang2023temporal,
  title={Temporal sentence grounding in videos: A survey and future directions},
  author={Zhang, Hao and Sun, Aixin and Jing, Wei and Zhou, Joey Tianyi},
  journal={IEEE Transactions on Pattern Analysis and Machine Intelligence},
  volume={45},
  number={8},
  pages={10443--10465},
  year={2023},
  publisher={IEEE}
}

@inproceedings{rao2025towards,
  title={Towards universal soccer video understanding},
  author={Rao, Jiayuan and Wu, Haoning and Jiang, Hao and Zhang, Ya and Wang, Yanfeng and Xie, Weidi},
  booktitle={Proceedings of the Computer Vision and Pattern Recognition Conference},
  pages={8384--8394},
  year={2025}
}

@inproceedings{kumar2025aligning,
  title={Aligning moments in time using video queries},
  author={Kumar, Yogesh and Agarwal, Uday and Gupta, Manish and Mishra, Anand},
  booktitle={Proceedings of the IEEE/CVF International Conference on Computer Vision},
  pages={20215--20225},
  year={2025}
}

@misc{statsbomb_opendata,
  author       = {{StatsBomb}},
  title        = {{StatsBomb Open Data}},
  year         = {2018},
  howpublished = {\url{https://github.com/statsbomb/open-data}},
  note         = {Accessed: 2025}
}

@article{hu2022lora,
  title={Lora: Low-rank adaptation of large language models.},
  author={Hu, Edward J and Shen, Yelong and Wallis, Phillip and Allen-Zhu, Zeyuan and Li, Yuanzhi and Wang, Shean and Wang, Liang and Chen, Weizhu and others},
  journal={Iclr},
  volume={1},
  number={2},
  pages={3},
  year={2022}
}

@inproceedings{carion2020end,
  title={End-to-end object detection with transformers},
  author={Carion, Nicolas and Massa, Francisco and Synnaeve, Gabriel and Usunier, Nicolas and Kirillov, Alexander and Zagoruyko, Sergey},
  booktitle={European conference on computer vision},
  pages={213--229},
  year={2020},
  organization={Springer}
}

@article{fawcett2006introduction,
  title={An introduction to ROC analysis},
  author={Fawcett, Tom},
  journal={Pattern recognition letters},
  volume={27},
  number={8},
  pages={861--874},
  year={2006},
  publisher={Elsevier}
}

@inproceedings{radford2021learning,
    title={Learning Transferable Visual
  Models from Natural Language Supervision},
    author={Radford, Alec and Kim, Jong Wook
  and Hallacy, Chris and Ramesh, Aditya and
  Goh, Gabriel and Agarwal, Sandhini and
  Sastry, Girish and Askell, Amanda and
  Mishkin, Pamela and Clark, Jack and
  Krueger, Gretchen and Sutskever, Ilya},
    booktitle={International Conference on
  Machine Learning},
    pages={8748--8763},
    year={2021}
  }

@inproceedings{feichtenhofer2019slowfast,
    title={SlowFast Networks for Video
  Recognition},
    author={Feichtenhofer, Christoph and Fan,
   Haoqi and Malik, Jitendra and He,
  Kaiming},
    booktitle={Proceedings of the IEEE/CVF
  International Conference on Computer
  Vision},
    pages={6202--6211},
    year={2019}
  }

@inproceedings{shao2020finegym,
  title={FineGym: A Hierarchical Video Dataset for Fine-grained Action Understanding},
  author={Shao, Dian and Zhao, Yue and Dai, Bo and Lin, Dahua},
  booktitle={Proceedings of the IEEE/CVF Conference on Computer Vision and Pattern Recognition (CVPR)},
  year={2020},
}
\end{document}